\documentclass[10pt,twocolumn,letterpaper]{article}

\usepackage{wacv}              

\usepackage{graphicx}
\usepackage{booktabs}
\usepackage{enumitem}
\usepackage{subcaption}
\usepackage{multirow}
\usepackage{amsfonts}
\usepackage{amssymb}
\usepackage{amsmath}
\usepackage{pifont}
\usepackage{bbding}
\usepackage{bbm}
\newcommand{\cmark}{\ding{52}}
\newcommand{\xmark}{\textcolor[rgb]{ .502,  .502,  .502}{\ding{55}}}
\usepackage{tcolorbox}
\usepackage{balance}
\usepackage{indentfirst}
\usepackage{makecell}
\usepackage{float}
\usepackage{ulem}
\usepackage{longtable}
\usepackage{supertabular}
\usepackage{hyperref}
\usepackage{silence}

\usepackage[capitalize]{cleveref}
\crefname{section}{Sec.}{Secs.}
\Crefname{section}{Section}{Sections}
\Crefname{table}{Table}{Tables}
\crefname{table}{Tab.}{Tabs.}

\newcommand{\boldparagraph}[1]{\vspace{0.0cm}\noindent{\bf #1}}

\def\wacvPaperID{431} 
\def\confName{WACV}
\def\confYear{2025}

\begin{document}

\title{MLLM-Tool: A Multimodal Large Language Model For Tool Agent Learning}

\author{Chenyu Wang$^{1}$,~
Weixin Luo$^{2}$,~
Sixun Dong$^{1}$,~
Xiaohua Xuan$^{3}$,~ \\
Zhengxin Li$^{1}$,~
Lin Ma$^{2}$,~ 
Shenghua Gao$^{4}$~ \\
$^1$ShanghaiTech University,
$^2$Meituan,
$^3$UniDT Technology,
$^4$University of Hong Kong \\
}
\maketitle

\begin{abstract}
Recently, the astonishing performance of large language models (LLMs) in natural language comprehension and generation tasks triggered lots of exploration of using them as central controllers to build agent systems. Multiple studies focus on bridging the LLMs to external tools to extend the application scenarios. However, the current LLMs' ability to perceive tool use is limited to a single text query, which may result in ambiguity in understanding the users' real intentions. LLMs are expected to eliminate that by perceiving the information in the visual- or auditory-grounded instructions. Therefore, in this paper, we propose MLLM-Tool, a system incorporating open-source LLMs and multi-modal encoders so that the learned LLMs can be conscious of multi-modal input instruction and then select the function-matched tool correctly. To facilitate the evaluation of the model's capability, we collect a dataset featuring multi-modal input tools from HuggingFace. Another essential feature of our dataset is that it also contains multiple potential choices for the same instruction due to the existence of identical functions and synonymous functions, which provides more potential solutions for the same query. The experiments reveal that our MLLM-Tool is capable of recommending appropriate tools for multi-modal instructions. Codes and data are available at \href{https://github.com/MLLM-Tool/MLLM-Tool}{github.com/MLLM-Tool/MLLM-Tool}.
\end{abstract}

\vspace{-10pt}

\section{Introduction}
\vspace{-5pt}
\label{sec:intro}
Large Language Models (LLMs) ~\cite{brown2020language, ouyang2022training, scao2022bloom, touvron2023llama} have showcased remarkable success in conversational scenarios.

With the promising capability of the instruction following via reinforcement learning from human feedback (RLHF) ~\cite{bai2022training, casper2023open, lee2023rlaif}, developers extend LLMs, e.g., ChatGPT with a Function Call feature, to the tool-use domain. This extension of LLMs' capabilities can be applied to a broad spectrum of situations, from intriguing outer tasks like restaurant reservations and online shopping to inner operations involving the retrieval and processing of infinite external knowledge. Along this path, numerous intelligent agents ~\cite{qin2023toolllm, tang2023toolalpaca,ruan2023tptu,yang2023autogpt, bubeck2023sparks, zeng2023glm130b, du2022glm, nakajima2023task} have been developed, by leveraging LLMs to accomplish various tasks in human-like ways.

\begin{figure}[t]
	\centering
	\includegraphics[width=1.0
 \linewidth]{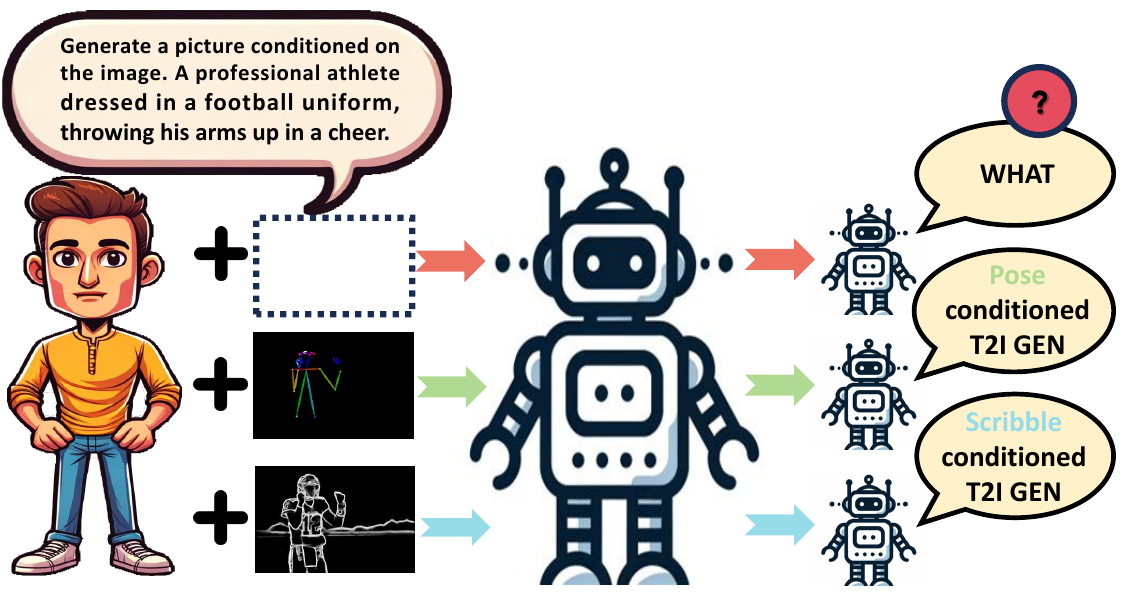}

	\caption{Comparisons of results outputted by tool agent with different inputs. The individual posed the same question thrice. Initially, no image was provided, causing the confusion. In the subsequent instances, images from varied scenes were uploaded, and our MLLM-Tool can correctly choose the correct tool based on the multi-modal inputs and generate distinct responses based on the differing pictorial contexts.}
	\label{fig:teaser}
 \vspace{-15pt}
\end{figure}

Although some efforts ~\cite{li2023api,xu2023tool,qin2023toolllm,patil2023gorilla} have been made to stimulate the capability to use external tools with LLMs, they are confined to purely textual inputs, leading to potential confusion and misinterpretation, especially when text is ambiguous in specifying the users' real intentions. As illustrated in Figure \ref{fig:teaser}, for the given textual prompt, \textit{``Generate a picture conditioned on the image. A professional athlete dressed in a football uniform, throwing his arms up in a cheer."}, LLMs struggle due to a lack of visual input and cannot specify the type of conditioned image. Recognizing this as an image generation task, they falter in recommending suitable pose- or scribble-conditioned text-to-image tools. However, when endowed with the capability to perceive visual or auditory instructions, LLMs, in this instance, can extract key \textit{``human pose"} and \textit{``scribble"} information from the second time and third time input images, respectively, thus resolving the ambiguity issue. Subsequently, multimodal instruction sensing is necessary to identify the appropriate tools for different tasks.

\begin{table}[t]
\centering
\renewcommand\arraystretch{1.0}
\setlength{\tabcolsep}{1pt}
\scalebox{0.85}{
\begin{tabular}{cccccc}
\toprule
Method                  & \makecell{Trainable \\LLM}          & \makecell{Multimodal \\LLM} & \makecell{Num. of\\modality}  & Multi-options \\ 
\midrule
MetaTool~\cite{huang2023metatool} & \xmark & \xmark & 1 & \xmark\\
APIBank~\cite{li2023api}  & \cmark & \xmark & 1 & \xmark\\
ToolBench~\cite{xu2023tool}  & \cmark & \xmark & 1 & \xmark\\
ToolLLM~\cite{qin2023toolllm}  & \cmark & \xmark & 1 & \xmark\\
APIBench~\cite{patil2023gorilla}  & \cmark & \xmark & 1 & \xmark\\
\hline
Visual ChatGPT~\cite{VisualChatGPT}  & \xmark & \xmark & 2 & \xmark\\
GPT4Tool~\cite{yang2023gpt4tools}  & \cmark & \xmark & 2 & \xmark\\
HuggingGPT~\cite{shen2023hugginggpt}  & \xmark & \xmark & 4 & \xmark\\
\hline
Ours  & \cmark & \cmark & 4 & \cmark\\ \bottomrule
\end{tabular}
}

\caption{Comparison of previous works and our work. The first five are intended for tool-augmented LLMs with textual inputs, and the middle three focus on empowering the task planning of LLMs and executing multi-modal inputs with corresponding tasks' APIs.}
\label{tab:benchmark}
\end{table}

\begin{figure}
\vspace{-10pt}
\centering{
  \includegraphics[width=1.0\linewidth]{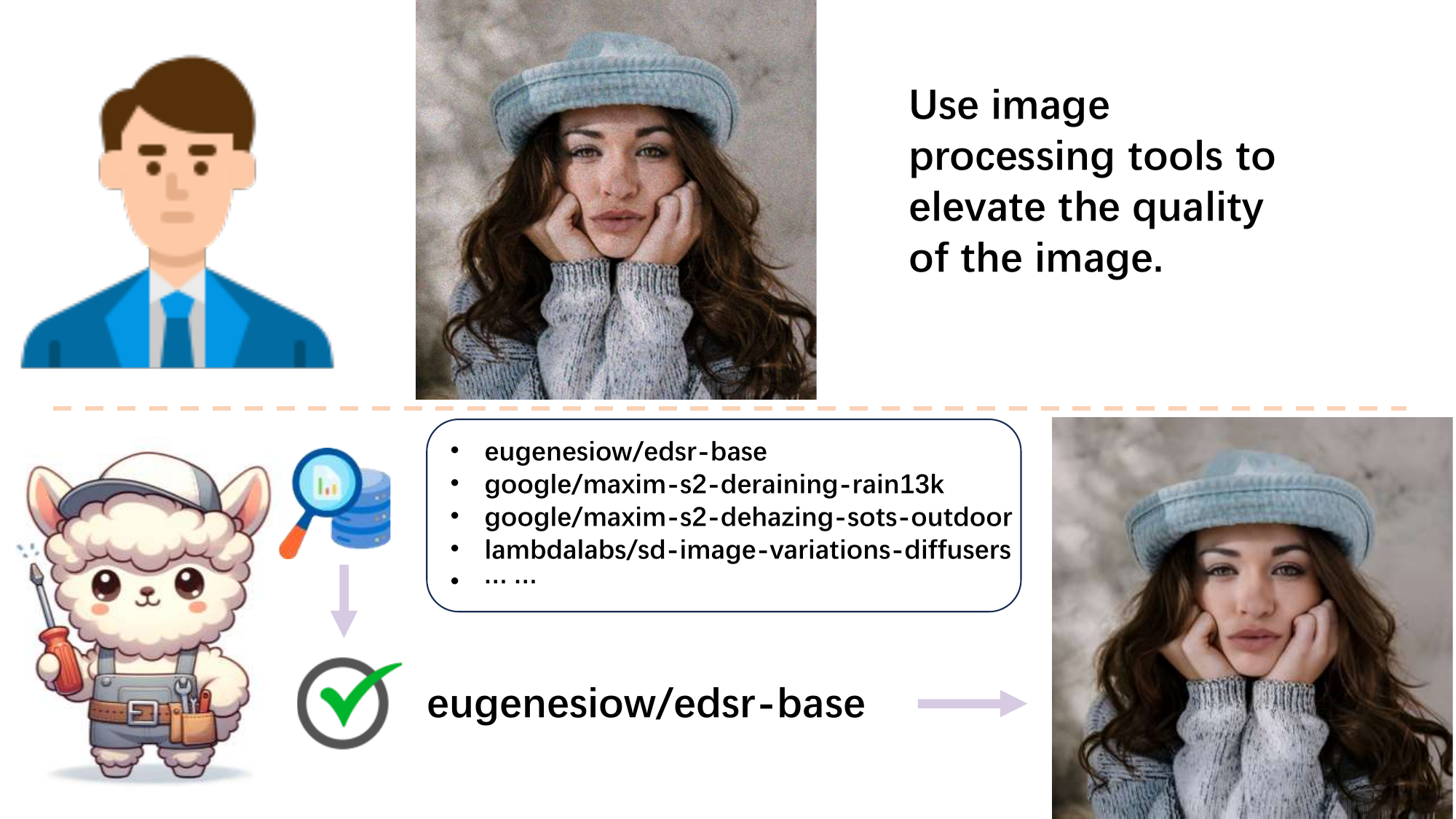}
}
\vspace{-5pt}
\caption{Qualitative example of MLLM-Tool. MLLM-Tool first receives the text instruction and perceives information from the input image to determine the corresponding task, selects the appropriate API that meets the requirements from the corpus, calls the API to execute, and finally gets the quality-improved image.}
\vspace{-12pt}
\label{fig:vis}
\end{figure}

\vspace{-1pt}
To remedy the previous works only accepting single text instructions, in this paper, we introduce a novel system, MLLM-Tool, integrating multi-modal encoders with open-source LLMs to synthesize multi-modal information for correct external tool identification. Table \ref{tab:benchmark} delineates the distinction between our work and previous works. In light of the fact that existing tool learning benchmarks ~\cite{li2023api,huang2023metatool,xu2023tool,qin2023toolllm,patil2023gorilla,VisualChatGPT,yang2023gpt4tools, shen2023hugginggpt}, do not explicitly construct text-to-any-modality pairs for alleviating text ambiguity, which cannot facilitate the training of our MLLM-Tool, in this paper, we collect a novel dataset, named ToolMMBench from the HuggingFace platform, one of the largest machine learning communities containing thousands of APIs. It is also worth noting that our ToolMMBench supports multi-options for each query, as demonstrated in Table \ref{tab:benchmark}. This feature is more closely related to real-life scenarios, where multiple alternatives exist to address the same challenges, albeit with varying performance levels. Such a feature would also make our dataset suitable for other applications in the future, such as the study of integrating multiple outputs from different tools to provide better feedback to the user. The process of our MLLM-Tool is shown in Figure \ref{fig:vis}.

In summary, our contributions are as follows:

\vspace{-6pt}
\begin{itemize}
    \item We develop MLLM-Tool, a multi-modal tool agent system that can perceive the visual- or audio-grounded instructions' information via incorporating multi-modal encoders and open-source LLMs.
    \vspace{-10pt}
    \item We construct a multi-modal input benchmark for evaluating LLMs' awareness and selection ability of the external tool usage with text-ambiguous queries. The benchmark offers content-aware multi-modal instructions and one-to-many instruction-answer pairs.
    \vspace{-10pt}
    \item  We design some evaluation metrics and conduct experiments with extensive subset ablations on multiple popular LLMs, and by fine-tuning MLLM-Tool on ToolMMBench, we observe a promising accuracy, 88.19\% of the tool selection, which demonstrates the effectiveness of our method.
\end{itemize}
\vspace{-10pt}

\section{Related Works}
\label{sec:related_works}

\textbf{Multi-modal Large Language Models.} LLMs have increasingly become the cornerstone of many natural language processing (NLP) applications. Subsequently, many efforts include GPT-4V ~\cite{GPT-4V}, LLaVA ~\cite{liu2023visual}, BLIP2 ~\cite{li2023blip}, Minigpt-4 ~\cite{zhu2023minigpt} developed the LLM to vision space by aligning off-the-shelf frozen pretrained image encoders. And similar ideas are generalized to video and audio modalities, like NeXTGPT ~\cite{wu2023next} and VideoLLaMA ~\cite{zhang2023video}. In contrast, our proposed MLLM-Tool focuses on exploring tool usage awareness and selection. Our goal is to empower the LLM's ability to be conscious of the external tools' function and select the suitable tools to act as a controller instead of enhancing the multi-modal understanding and reasoning ability of MLLMs.  

\textbf{Tool Agent.} Recent advancements have witnessed a rapid emergence of systems that deploy LLMs as a central orchestrator for tool assistance. Toolformer ~\cite{schick2023toolformer} pioneers the study of bridging the LLMs with external tools. Subsequent works have expanded the application domains of this integration to encompass health support ~\cite{ma2023understanding}, code synthesis ~\cite{li2023api,suris2023vipergpt}, web searching ~\cite{song2023restgpt}. Gorilla ~\cite{patil2023gorilla} collected a machine learning domain dataset and introduced API retrieval to boost the performance. ToolLLM ~\cite{qin2023toolllm} focused on executing complex tasks in practical scenarios. GPT4Tools ~\cite{yang2023gpt4tools} and Visual ChatGPT ~\cite{VisualChatGPT} integrated visual found models after decomposing the task. It is worthwhile to note that although previous works, like HuggingGPT ~\cite{shen2023hugginggpt}, also handle multimodal data, the non-text data is input in a URL format. After task planning based on the text information, the selected models process the data, so the non-text data are not directly perceived. On the contrary, we concentrate on relying entirely on the ability of large language models to naturally process multimodal inputs and select the appropriate APIs, which makes the agent smart enough to understand the user’s intentions and avoid ambiguity.
\section{Dataset Construction}
\label{sec:dataset}

In this section, we present the process of constructing a benchmark from HuggingFace, which primarily involves three steps: collecting API datasets (\cref{sec:api}), generating instruction-answer pairs (\cref{sec:pairs}), and constructing one-to-many instruction matching (\cref{sec:matching}). We also introduce the statistics and distribution of the benchmark in \cref{sec:stat}.

\subsection{API Collection}\label{sec:api}

Initially, we crawled 263,945 models from the HuggingFace platform. However, the quality of these models varies significantly, for example, some models are poorly maintained and even lack information in the model card. To ensure only those models meeting specific criteria (e.g., having a detailed model card, especially providing a model description and example script) are ultimately selected, we employ a meticulous filtering process, and the detailed filtering process is shown in the Appendix. Finally, we only retain 932 high-quality models. 

We document a detailed model card for each model inspired by ~\cite{patil2023gorilla}. The model card includes essential information such as domain, API name, API call, parameters, example codes, and descriptions. Further, to prevent an overabundance of APIs with identical functionalities leading to excessive options for subsequent instruction matching (e.g., in the imaging realm, where two APIs might differ only in their network architectures but are identical in terms of the datasets they were trained on and their functionalities), we empirically set a maximum of five APIs per functionality category, prioritizing the most downloaded ones to streamline the dataset while maintaining diversity. This limit is informed by our experience with API distributions across tasks and practical considerations for subsequent evaluations. Specifically, some of our metrics, explained in detail later, require multiple repeated inferences to assess the proportion of predicted APIs that match ground truth APIs. Increasing the number of APIs per category would significantly increase computational resource requirements.

It is worth noticing that the function classification rules formulated by the HuggingFace platform are coarse-grained and not suitable for direct use by our work for some reasons. In particular, some models lack assigned categories, are incorrectly classified, or contain multiple tasks, but only one is labeled. Additionally, the platform employs overly coarse, resulting in a loss of fine-grained distinctions. For example, the task of \textit{Audio Classification} can be subdivided into \textit{Event Recognition, Command Recognition, Spoken Language Identification, Speaker Verification, \textit{etc}}. We then propose a new hierarchical function categorization system and rearrange the filtered model. We list all the tasks our system supports and the subtasks we divide under these tasks in the Appendix. The corresponding supported tasks of the model are stored under the keys coarse functionality and fine functionality, which are then integrated into the corresponding model card. Given that our API collection may not cover all tasks, users in real-world scenarios might encounter situations without a suitable API. Thus, we also introduce the \textit{``Unknown"} option to address this issue.

\begin{figure}[htbp]
	\centering
	\includegraphics[width=0.9\linewidth]{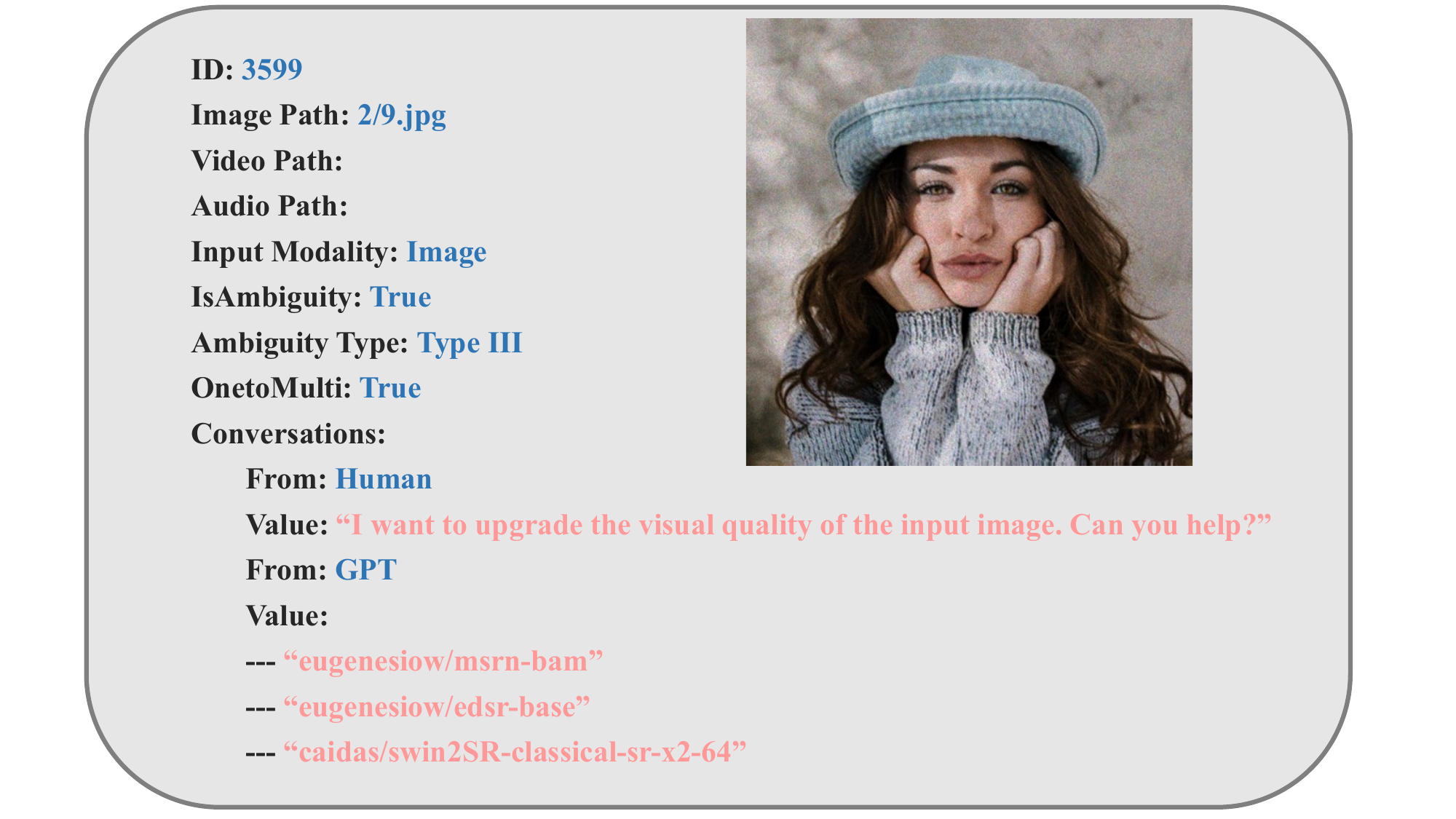}
	\caption{Visualization of sampled data in the dataset. We record the input modality information and check for the presence of text ambiguity; if ambiguity exists, we further document the type of ambiguity by its index and note whether multiple APIs can resolve the query.}
	\label{fig:dataset_sample}
        \vspace{-15pt}
\end{figure}

\subsection{Instruction-Answer Pairs Generation}\label{sec:pairs}
To curate the instruction-answer pairs, we leverage GPT-4 to understand the functionalities of these API calls and then generate 20 queries for each API call. Detailed prompts are provided in the Appendix.

After meticulously filtering and refining through manual review, 10 well-crafted queries will be finally selected to be associated with each API call. A significant highlight of our approach is its capability to disambiguate text prompts by learning the characteristics of other modalities when handling multi-modal inputs. With the assistance of GPT, We elaborate on five text ambiguity cases that need to be checked:

\textbf{Data models used are from different domains}:  It is imperative to acknowledge the inherent disparities across data domains, either auditory or visual. Owing to these pronounced differences between domains, a unified treatment could result in significant biases in outcomes. Models need to judiciously select APIs contingent on the particular data domain at hand, instead of using a unified treatment. For instance, in image segmentation tasks, selecting distinct APIs tailored for specific scenarios like indoor decorations, portraits' fashion apparel, or medical related according to the given image is vital.

\textbf{Distribution and granularity of categories in different datasets}: 
Within the same domain, we need to consider the covered categories and category granularity variations. Especially when doing image classification tasks, specific strategies vary for common, specialized, or rare categories.

\textbf{Image Quality}: The most evident instance revolves around image quality, which can be compromised by multiple factors, including fog, rain, low resolution, and motion blur. When confronted with vague commands such as \textit{``I want to enhance the image quality"}, analyzing the reason for the low imaging quality of the input image and then selecting the appropriate API call for quality enhancement becomes necessary.

\textbf{Input Conditions of different models}: In some cases, the format of the input image can also assist in resolving text ambiguity issues. For example, ControlNet ~\cite{zhang2023adding} posits multiple conditions for image generation. Faced with an instruction like \textit{``Based on the provided image, generate a photo of a chef in the kitchen,"} and given inputs such as a posed image or a normal map, the model must leverage the unique features distinguishing between the two types of images. This discernment facilitates the selection of the appropriate API to achieve the desired output.

\textbf{Others}: There are some individual cases unique to individual modality, for instance, linguistic variations in the audio realm. Within the speech recognition task, when presented with an instruction like \textit{``In my language class, we listened to a story. I managed to record it; can you tell me its content?"}, the absence of explicitly provided language information necessitates the model first to ascertain the spoken language. Only then can it aptly select the corresponding API for accurate content extraction and interpretation.

For convenience, the following defaults to five ambiguity situations, referred to as \textbf{Domains, Categories, Quality, Conditions and Others}.

Subsequently, nuanced manual intervention is warranted to search for prompt-matching inputs,  especially in scenarios involving multi-modalities, sourced either from relevant datasets or online searches. Take the image classification task for example, one might say, \textit{``I have a photo of marine life taken during my scuba diving trip. Can you identify its category?"}, it becomes crucial to provide a marine creature image input to match the context of the prompt \textit{``marine life"} and \textit{``scuba diving trip"}, and the corresponding image path would be saved.

We collect a total of multi-modal instruction-answer pairs, including 3798 image inputs, 103 video inputs, and 726 audio inputs, totaling 72 videos, 3777 images, and 590 audio files. Details about the annotation effort and non-text data collection process are shown in the Appendix.

\subsection{Instruction Matching}
\label{sec:matching}
In contrast to prior works, our dataset stands out with a unique feature: it encompasses numerous API calls that perform identical or similar functions. This results in a scenario where a single instruction could be associated with several API calls within our dataset, each potentially suitable for addressing it.
In step two, we approach from a singular API function perspective to construct one-to-one matched instruction-answer pairs, now necessitating integration of results from the classification system established in the first step. For APIs with identical functionalities, we can match their queries immediately. However, for groups of APIs with similar functionalities, it is essential to determine whether a query aligns with a strict or broad functional description of each API. For instance, the request, \textit{``Can you help me generate an image of a 'golden-haired magical princess' in the Disney style?"} In this scenario, the emphasis on the \textit{``Disney style"} narrows the choice of APIs to those pre-trained on Disney-themed images. Conversely, a more general instruction like, \textit{``I'm organizing a role-playing event and need a reference image. Can you generate an image of a 'masked mysterious hero' for me?"} leaves room for multiple APIs to be potential solutions, given its lack of specificity. This process yielded a dataset where each instruction corresponds to one or, in some cases, multiple relevant API calls. 

\begin{table}[htbp]
\resizebox{1\columnwidth}{!}{
    \begin{tabular}{c|c|c|c|c}
    \hline
    Settings  & Text & Audio & Image & Video \\
    \hline
    Training Set (w/ split) & 31,280  & 2,944  & 10,085 & 224  \\
    \hline
    Training Set (w/o split) & 5,563 & 562  & 3,938  & 82  \\
    \hline
    Testing Set & 1,452  & 164  & 760 & 21  \\
    \hline
    \end{tabular}}%
    \centering
  \caption{Distribution of training set and testing set. Note that, in the training set, one-to-many scenarios are split into multiple one-to-one instructions, so the first line represents the number of queries after splitting while the second line calculates the distinct query number of four single modalities. The last line is the distribution of the whole testing set.}
  \label{tab:dataset0}%
  \vspace{-10pt}
\end{table}%

\subsection{Dataset Statistics}\label{sec:stat}
In alignment with HuggingFace's task-specific API distributions, ToolMMBench exhibits a similar long-tailed distribution, with text-related APIs being the most abundant and video-related ones the least. ToolMMBench consists of 27 coarse-grained task categories inheriting from HuggingFace, from which we expand to 29 tasks to form the first level of our classification system and then extend to our predefined fine-grained tasks. 

\begin{figure*}[htbp]
	\centering
	\includegraphics[width=0.96\linewidth]{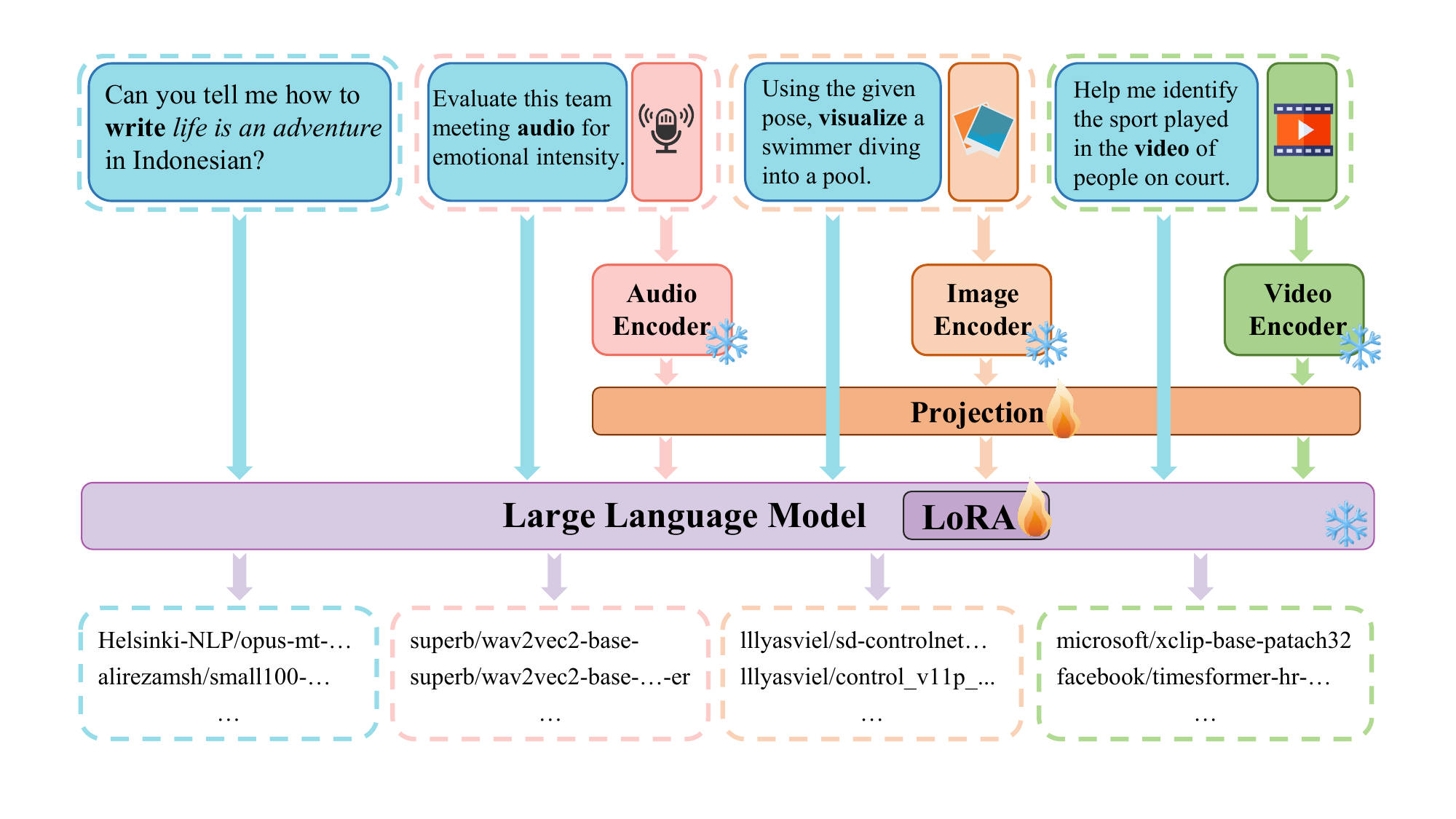}
    \vspace{-30pt}
	\caption{\textbf{Our network architecture}. {The network could process four distinct types of input, each uniquely color-coded. These inputs, irrespective of their modality, are first encoded using modality-specific frozen encoders. Subsequently, they are passed through a trainable projection layer, aligning them within a unified language feature space. The fine-tuned Large Language Models (LLMs), augmented with Low-Rank Adaptation (LoRA), then process a combination of the prompt and the projected embeddings, enabling the accurate prediction of the corresponding API name.} }
	\label{fig:network}
    \vspace{-12pt}
\end{figure*}


We divide our dataset into training and testing sets in an 8:2 ratio. Since our task currently focuses on a closed set of APIs, our test set cannot include any APIs that have not been encountered in the training set. Therefore, we approach this from the perspective of queries generated by each API. Given the presence of one-to-many query scenarios in our dataset, we implement a strategy in the training set to address these cases. Specifically, we split each one-to-many query into \textit{W} distinct inputs, where \textit{W} represents the number of APIs that support the given query. This approach ensures that each potential API response is individually represented, allowing the model to learn the diverse range of suitable responses for a single query. In contrast to the training sampling strategy, for the one-to-many query scenarios in our dataset, each query appears only once in the test set. Finally, the training set comprises 44,533 instructions, containing 9,245 different instructions, and the testing set totaled 2,397 instructions. The detailed distribution is illustrated in Table \ref{tab:dataset0}. Figure \ref{fig:dataset_sample} displays a data sample from the dataset. The visualized model cards and additional data samples are available in the Appendix.

\section{MLLM-Tool}\label{sec:method}

We introduce MLLM-Tool, a method that leverages the LLM to perform the role of a multi-modality tool agent. The architecture of this network is depicted in Figure \ref{fig:network}. Considering the diversity of our data, which includes images, videos, and audios, we incorporate ImageBind ~\cite{girdhar2023imagebind} as our primary multimodal encoder. ImageBind ~\cite{girdhar2023imagebind} is designed to create a unified embedding space that integrates six distinct modalities, utilizing the advancements in large-scale vision-language models.

For the large language model component of MLLM-Tool, we select a range of leading LLMs, including Vicuna ~\cite{vicuna}, Llama ~\cite{touvron2023llama}, Llama2 ~\cite{touvron2023llama2}, and Llama2-Chat ~\cite{touvron2023llama2}. These models are chosen for their exemplary performance in various areas, such as inference capabilities, knowledge retention, and comprehension skills. Consequently, our system is equipped with the proficiency to interpret multi-modal input instructions effectively.

In our dataset, there are four types of modality-combination input: \textit{Text}, \textit{Text+Image}, \textit{Text+Video}, and \textit{Text+Audio}. The output from the LLM includes textual responses that provide tool recommendations along with their respective model card information. For the multimodal inputs in the \textit{Image}, \textit{Video}, or \textit{Audio} modalities, we first utilize ImageBind ~\cite{girdhar2023imagebind} encoders with fixed weights for feature extraction. These features are then aligned with the LLM's feature space through a linear projection layer. To optimize our model while minimizing the number of learnable parameters, we implement Low-Rank Adaptation (LoRA) ~\cite{hu2021lora} to fine-tune an open-source LLM.

\vspace{-5pt}

\section{Evaluation Metrics}\label{metrics}

Since no prior work addresses user prompt ambiguity from the perspective of constructing text-to-any-modality pairs, there are no established metrics for measuring performance. In this paper, we propose evaluation metrics tailored to the dataset's unique attributes, ambiguity types, multiple options, and diverse modality inputs to effectively measure performance disparities.

\textbf{Ambiguity types}. Since we systematically document whether each case involves ambiguity and record the corresponding ambiguity type index during the annotation process, as exemplified by the cases shown in Figure \ref{fig:dataset_sample}, we construct the corresponding testing subsets based on the five distinct ambiguity scenarios delineated above. These facilitate the evaluation of the model's capability to leverage multimodal information under varying types of ambiguity, providing a nuanced understanding of its performance.
    
\textbf{Multiple options}. Since our model has the capacity to support multiple options, we divide the instruction set into two subsets, solved only by one single API versus those allowing for multiple API options, and we expect to examine the performance gaps within two subsets.

\textbf{Individual modality}. For each modality, we partition subsets to evaluate model performance with fixed modality combinations, ensuring a comprehensive assessment across varying input configurations.

The testing set is divided into several subsets based on the three attributes. For a subset $\mathbf{S}$, supposing containing $n$ instructions $\{x_{1}, x_{2} \dots x_{n}\}$, where each instruction $x_{i}$ matches an API list $y_{i}$, which consists of a list of $M$ suitable APIs $[A_{i}^{1},A_{i}^{2}, \dots, A_{i}^{M}]$. Thus, the collection of API lists corresponding to this instruction subset combined to a set $\mathbf{Y}$, $\{y_{1}, y_{2} \dots y_{n}\}$. With these symbols defined, assuming the subset $\mathbf{S}$ yields an API set $\{z_{1}, z_{2} \dots z_{n}\}$, output from our model, we can represent the accuracy of each mentioned subset using the following formula: $\text{Acc} = \sum \limits_{i=1}^{n} \mathbbm{1}_{z_{i} \in y_{i}} / n$
\vspace{-1pt}

The above formula also applies to the accuracy calculation for the entire testing set. Besides that, the hallucination rate is used as a metric to evaluate the performance of each testing subset and the overall testing set. To calculate the hallucination rate, we first assume the corpus $\textsc{C}$, containing all APIs. Hallucination refers to the phenomenon where the model outputs APIs are not present in the corpus $\textsc{C}$, essentially fabrications of the model. Taking the above subset $\mathbf{S}$ as an example, the calculation of the hallucination rate can be expressed with a formula:  $\text{Hallu} = \sum \limits_{i=1}^{n} \mathbbm{1}_{z_{i} \notin C} / n$.
\vspace{-2pt}

Besides that, unlike simple one-to-one mapping relationships, using single-instance accuracy is insufficient for evaluating the system's comprehension of the relationships of APIs, especially in one-to-many mapping relationships. Hence, we propose using recall for repeated inferences. By examining the proportion of ground truth encompassed by the collective results of multiple inference results, marked as $R_{i}^{j}$, where $i$ is the $i^{th}$ instruction in the testing set and $j$ represents for the $j^{th}$ inference. The corresponding metric calculation is expressed as follows: 
\begin{equation*}
\begin{small}
    \text{Recall} = \frac{1}{n} \sum_{i=1}^{n} \frac{(f_{c}(Set({R_{i}^{1}, R_{i}^{2} \dots R_{i}^{K} |R_{i}^{j} \in y_{i}, j \in [1,K]}))}{f_{c}(y_{i})}
\end{small}
\vspace{-7pt}
\end{equation*}

where $K$ is the inference times, $f_{c}$ is a counting function.

Additionally, to assess whether the system's output is intuitive and whether the output format meets user needs, we introduce a metric named \textbf{format accuracy}. Specifically, we aim for the output to be presented in JSON format, with clearly delineated key-value pairs. In this paper, the format should be \textit{\{api\_name: [Recommended API]\}}. Any alterations to this format, including missing symbols or changing the order of placement, are all considered formatting errors. This structure is designed to facilitate user navigation, allowing for easy selection of desired sections. 

\vspace{-5pt}

\section{Experiments}
In this section, we conduct experiments on our collected dataset, evaluating the performance of our system MLLM-Tool. We start by introducing the experiment setup in \cref{setup}. In \cref{main}, we assess MLLM-Tool's ability on our benchmark, followed by three specific conditions’ performance analysis in \cref{analysis}. We ablate on the encoders to validate the effectiveness in \cref{ablation}. One visualized result is shown in Figure \ref{fig:vis}, and more are shown in the Appendix.

\subsection{Experimental Setup}\label{setup}
The setup follows ~\cite{su2023pandagpt} and adjusts based on our computation resources (4$\times$A40 GPUs). The more detailed setup and the data processing method are shown in the Appendix.

\begin{table}[htbp]
\renewcommand{\arraystretch}{0.7}
\resizebox{0.9\columnwidth}{!}{
    \begin{tabular}{cccc}
    \hline
    Model  & Acc $\uparrow$ & Hallu $\downarrow$  & Format Acc$\uparrow$ \\
    \hline

    Vicuna-7B  & 83.15  & 1.63  & 100.00  \\
    Llama-7B  & 84.48  & 0.58  & 100.00  \\
    Llama2-7B  & 80.68  & 1.96  & 100.00  \\
    Llama2-Chat-7B & {83.60}  & {1.34}  & 100.00  \\
    \hline

    Vicuna-13B  & \textbf{88.19}  & \uline{0.17} & 100.00\\
    Llama-13B & \uline{87.86}  & \uline{0.17}  & 100.00\\
    Llama2-13B & 81.81  & \uline{0.17}  & 100.00\\
    Llama2-Chat-13B & 87.78  & \textbf{0.08} & 100.00\\
    \hline
    \end{tabular}}%
    \centering
  \caption{\textbf{Experimental results of MLLM-Tool.} We present the performance of four types of LLMs with different configurations.}
  \label{tab:suppl_combined}%
\end{table}%

\vspace{-10pt}

\subsection{Evaluation of Fine-Tuned LLMs and Comparative Performance}\label{main}
We finetune four open-source LLMs, Llama, Llama2, Llama2-Chat, and Vicuna, each with two 7B and 13B variants, on the instruction-solution data pairs. As illustrated in Table \ref{tab:suppl_combined}, we report the accuracy, hallucination rate, and format accuracy of the whole test dataset, which are detailed in Sec. \ref{metrics}. Due to the limitation of computing resources, for the recall metric, we focus on two models, Llama2-7B, and Llama2-Chat-7B, setting the inference times $K$ to 10. They achieve recall rates of 62.13\% and 61.47\%, respectively.

To compare with closed-source MLLMs, we select 80 cases with an equal number of text and image inputs. We choose GPT-4o-128K ~\cite{GPT-4V}, Claude-3-Opus-200K ~\cite{Claude_3}, and Gemini 1.5 Pro ~\cite{team2023gemini} for evaluation based on their performance and supported context length. The input integrates streamlined descriptions of all APIs with instructions. Table \ref{tab:close_llms} shows their zero-shot and few-shot performance.

\begin{table}[t!]
\begin{small}
\centering
  \renewcommand{\arraystretch}{0.8}
  \resizebox{0.9\columnwidth}{!}{
    \begin{tabular}{ccccccccccc}
    \hline
    \multirow{2}{*}{Model} & \multicolumn{2}{c}{Text} & \multicolumn{2}{c}{Image}\\
\cmidrule(lr){2-3} \cmidrule(lr){4-5} \cmidrule(lr){6-7}     
& Acc$\uparrow$ & \multicolumn{1}{c}{Hallu$\downarrow$} & Acc$\uparrow$ & Hallu$\downarrow$ \\
    \hline

    GPT (zero-shot) & 25.0 & 2.5 & 5.0 & 17.5 \\
    GPT (few-shot) & 20.0 & 5.0 & 25.0 & 20.0  \\
    Claude (zero-shot) & 45.0 & \textbf{0.0} & 17.5 & 5.0 \\
    Claude (few-shot) & 32.5 & 7.5 & 17.5 & 5.0  \\
    Gemini (zero-shot) & 22.5 & 2.5 & 32.5 & 5.0  \\
    Gemini (few-shot) & 35.0 & \textbf{0.0} & 22.5 & 10.0  \\ \hline
    Ours (Vicuna-13B) &  \textbf{87.5} & \textbf{0.0}  & \textbf{85.0} & \textbf{0.0} \\
    \hline
     \end{tabular}}%
\caption{Comparisons with close-sourced MLLMs. We select 40 cases each from text and image input modalities, and the results demonstrate that our model achieves a significant performance advantage in terms of accuracy over other models.}
\label{tab:close_llms}%
\vspace{-20pt}
\end{small}
\end{table}%

In comparisons with open-sourced MLLM, we select LLaVA-v1.5  ~\cite{liu2023visual}, which demonstrates outstanding performance in multimodal tasks. Given the constraints of input modality, we separate the image-input data and fine-tuned LLaVA-v1.5 accordingly. To ensure a fair evaluation, we retrain our MLLM-Tool model using only the same image-input data. The results are shown in Table \ref{tab:llava}. Note that, due to LLaVA's overall performance being inferior to that of closed-source MLLMs and its limited support for input token length, we do not include a comparison with LLaVA's zero-shot performance here.

We can derive some findings from the results:

1.Given the larger parameter size of large language models, accuracy improves, and the hallucination phenomenon becomes less frequent. Notably, Vicuna-13B achieves an accuracy of 88.19\%. When compared with closed-source MLLMs, the significant improvement underscores the necessity of fine-tuning open-source LLMs. Compared to open-source models, both models achieve strong performance at the same parameter scale, but our model architecture shows a slight advantage in terms of both accuracy and hallucination rate;
2.The format maintains a hundred percent accuracy across all models, demonstrating that LLMs have a high understanding of structural rules, which is crucial for tasks that demand strict formatting;
3.The hallucination rate generally remains low, which we speculate is due to the relative simplicity of our dataset. Since most of our data provides a clear context and our extracted description does not involve complex language structures so that the model is more likely to produce responses within the whole corpus; 
4.The overall recall values remain modest. We hypothesize this is due to an abundance of selectable APIs for some queries. Despite limiting identical function APIs, common object labels appearing across multiple datasets can accumulate suitable APIs in cases like image classification, potentially exceeding ten options. Therefore, even with ten inference times, it is impossible to cover all.

Overall, the results affirm that MLLM-Tool effectively brings out the tool-use potential in LLMs and enhances LLMs to deal with multi-modal instruction inputs.

\vspace{-5pt}
\subsection{Specific Conditions' Performance Analysis}\label{analysis}
To verify the effectiveness of our system when encountering different complex cases, we conduct experiments on three specific conditioned testing subsets, as defined in \cref{metrics}, to assess the impact of different data characteristics on our model's performance and accuracy. 

\begin{table}[htbp]
\centering
\renewcommand{\arraystretch}{0.8}
\resizebox{0.9\columnwidth}{!}{
    \begin{tabular}{cccc}
    \hline
      Method &  Acc  $\uparrow$ & Hallu $\downarrow$ & Format Acc $\uparrow$ \\
    \hline
     LLaVA-v1.5-7B  & 91.05 & 0.66 & 100.00 \\
     LLaVA-v1.5-13B & 93.15 & 0.52 & 100.00\\ \hline
     Ours(Vicuna-7B)&    \uline{93.68} & \textbf{0.00} & 100.00 \\
     Ours(Vicuna-13B) &  \textbf{94.44} & \uline{0.13} & 100.00\\
    \hline
    \end{tabular}}%
     \caption{Comparison with open-sourced MLLM. Both model architectures perform exceptionally well, with ours showing a slight advantage at the same parameter scale.}
    \label{tab:llava}
    \vspace{-10pt}
\end{table}

\begin{table*}[htbp]
\centering
\renewcommand{\arraystretch}{0.9}
\resizebox{2.0\columnwidth}{!}{
\begin{tabular}{c c c c c c c c c c c c c c c}
\hline
\multirow{2}[2]{*}{Model} & \multicolumn{2}{c}{Domains} & \multicolumn{2}{c}{Categories} & \multicolumn{2}{c}{Quality} & \multicolumn{2}{c}{Conditions} & \multicolumn{2}{c}{Others} & \multicolumn{2}{c}{With ambiguity} & \multicolumn{2}{c}{Without ambiguity} \\ 
\cmidrule(lr){2-3} \cmidrule(lr){4-5} \cmidrule(lr){6-7} \cmidrule(lr){8-9} \cmidrule(lr){10-11}\cmidrule(lr){12-13}\cmidrule(lr){14-15}
& \multicolumn{1}{c}{Acc$\uparrow$} & \multicolumn{1}{c}{Hallu$\downarrow$} & \multicolumn{1}{c}{Acc$\uparrow$} & \multicolumn{1}{c}{Hallu$\downarrow$} & \multicolumn{1}{c}{Acc$\uparrow$} & \multicolumn{1}{c}{Hallu$\downarrow$} & \multicolumn{1}{c}{Acc$\uparrow$} & \multicolumn{1}{c}{Hallu$\downarrow$} & \multicolumn{1}{c}{Acc$\uparrow$} & \multicolumn{1}{c}{Hallu$\downarrow$} & \multicolumn{1}{c}{Acc$\uparrow$} & \multicolumn{1}{c}{Hallu$\downarrow$} & \multicolumn{1}{c}{Acc$\uparrow$} & \multicolumn{1}{c}{Hallu$\downarrow$} \\ \hline


    Vicuna-7B & {83.91}  & 2.30  & 72.55  & 0.00  & \textbf{100.00}  & 0.00  & 94.25  & 0.00  & \textbf{100.00}  & 0.00  & {91.37}  & 0.94  & 86.64  & 0.46  \\
    Llama-7B & 75.86  & 0.00  & 66.67  & 0.00  & 73.53  & 0.00  & {95.21}  & 0.00  & \textbf{100.00}  & 0.00  & 88.37  & 0.00  & 85.94  & 0.46  \\
    Llama2-7B & 66.67  & 0.00  & 66.67  & 0.00  & 94.12  & 0.00  & 92.97  & 0.00  & \textbf{100.00}  & 0.00  & 86.87  & 0.00  & 83.50  & 0.72  \\
    Llama2-Chat-7B & 87.36  & 0.00  & {74.51}  & 0.00  & \uline{97.06}  & 0.00  & 95.85  & 0.00  & \textbf{100.00}  & 0.00  & 92.87  & 0.00  & {80.07}  & 3.41 \\

    \hline
    Vicuna-13B & 88.51  & 0.00  & 66.67  & 0.00  & \uline{97.06}  & 0.00  & \textbf{97.76}  & 0.00  & \textbf{100.00}  & 0.00  & 93.43  & 0.00  & \uline{86.91}  & 0.21 \\
    Llama-13B  & \uline{89.66}  & 0.00  & \uline{76.47}  & 0.00  & \textbf{100.00}  & 0.00  & 97.12  & 0.00  & \textbf{100.00}  & 0.00  & \uline{94.37}  & 0.00  & \textbf{89.31}  & 0.00  \\
    Llama2-13B & 64.37  & 0.00  & 60.78  & 0.00  & 73.53  & 0.00  & 73.48  & 0.32  & \textbf{100.00}  & 0.00  & 73.17  & 0.19  & 77.65  & 0.00  \\
    Llama2-Chat-13B & \textbf{90.80}  & 0.00  & \textbf{78.43}  & 0.00  & \uline{97.06}  & 0.00  & \uline{97.44}  & 0.00  & \textbf{100.00}  & 0.00  & \textbf{94.75}  & 0.00  & 84.19  & 0.00\\
    \hline

\end{tabular}}
\caption{The performance on testing subsets divided by ambiguity type. For the \textit{without ambiguity}, we only focus on non-text input data. Notably, for the ambiguity types \textit{Quality, Conditions, and Others}, the accuracy of selecting the correct API is noticeably higher.}
\label{tab:suppl_ambiguity}
\end{table*}%

\textbf{Ambiguity types}. In Table \ref{tab:suppl_ambiguity}, we present results for five types of ambiguity and compare non-ambiguity with ambiguity cases. To eliminate the influence of purely text-based data, which inherently lacks ambiguity, we focus solely on the performance of non-text input in the \textit{without ambiguity} scenario. We find that ambiguity types I (Domains) and II (Categories) perform worst in handling ambiguity, while other categories, especially type V (Others), achieve 100\% accuracy across all models. The strong performance of types III (Quality) and IV (Conditions) likely stems from apparent visual differences and classification boundaries, while types I and II, especially II, face challenges due to varying dataset granularity. The model must understand the multimodal content and the appropriate classification granularity, complicating API selection. Surprisingly, performance without ambiguity falls short of that with ambiguity, highlighting that even detailed text descriptions cannot fully substitute for multimodal data. Detailed text can introduce noise, affecting model selection accuracy. This emphasizes the need for LLMs to process visual and auditory instructions, reinforcing the significance of our work.

\begin{table*}[t!]
\begin{footnotesize}
\centering
  \renewcommand{\arraystretch}{0.9}
    \begin{tabular}{ccccccccccccc}
    \hline
    \multirow{2}{*}{Model} & \multicolumn{2}{c}{one-to-one} & \multicolumn{2}{c}{one-to-many} & \multicolumn{2}{c}{Video} & \multicolumn{2}{c}{Audio} & \multicolumn{2}{c}{Image} & \multicolumn{2}{c}{Text}\\
\cmidrule(lr){2-3} \cmidrule(lr){4-5} \cmidrule(lr){6-7} \cmidrule(lr){8-9} \cmidrule(lr){10-11} \cmidrule(lr){12-13}     
& Acc$\uparrow$ & \multicolumn{1}{c}{Hallu$\downarrow$} & Acc$\uparrow$ & Hallu$\downarrow$ & Acc$\uparrow$ & Hallu$\downarrow$  & Acc$\uparrow$ & Hallu$\downarrow$  & Acc$\uparrow$ & Hallu$\downarrow$  & Acc$\uparrow$ & Hallu$\downarrow$\\
    \hline

    
    Vicuna-7B  & 66.03  & 3.35  & 90.01  & 0.94  & 71.43  & 0.00  & \textbf{93.29} & 0.00  & 88.95 & 0.53  & 79.13  & 2.41 \\
    Llama-7B & {69.24}  & 1.31  & 90.59 & 0.29  & 76.19  & 0.00  & 89.02  & 0.00  & {87.24}  & 0.26  & 82.64  & 0.83  \\
    Llama2-7B & 63.12  & 3.64  & 87.73  & 1.29  & \uline{80.95}  & 0.00  & \textbf{93.29} & 0.00  & 83.82  & 0.39  & 77.62  & 3.03 \\
    Llama2-Chat-7B & 70.99 & 2.33  & 88.66  & 0.94  & \uline{80.95} & 0.00  & \uline{92.07}  & 0.00  & 86.45  & 1.84  & {81.20}  & 1.24 \\
    \hline
    
    Vicuna-13B & \textbf{77.26}  & 0.58  & \textbf{92.58}  & 0.00 & 76.19  & 0.00  & \uline{92.07}  & 0.00  & \uline{90.66}  & 0.13  & \textbf{86.64}  & 0.21 \\
    Llama-13B & \uline{76.68}  & 0.58  & \uline{92.34}  & 0.00  & 71.43  & 0.00  & \uline{92.07}  & 0.00  & \textbf{92.76}  & 0.00  & 85.06  & 0.28  \\
    Llama2-13B & 69.24  & 0.44  & 86.85  & 0.06  & \textbf{85.71}  & 0.00  & 89.02  & 0.00  & 71.84  & 0.13  & 86.16  & 0.21 \\
    Llama2-Chat-13B & 76.38  & 0.29  & \uline{92.34}  & 0.00 & \uline{80.95}  & 0.00  & 89.63  & 0.00  & 90.53  & 0.00  & \uline{86.23}  & 0.14 \\
    \hline
     \end{tabular}%
\caption{It is observed that when multiple API options are available for a given query, the accuracy of selecting the correct API is higher. In terms of input modalities, the system performs best on queries with \textit{Audio} and \textit{Image} inputs, highlighting its strength in handling these data types.}
\label{tab:suppl_one2mul}%
\vspace{-12pt}
\end{footnotesize}
\end{table*}%

\vspace{-1pt}
\textbf{Multiple options}. We divide the testing set into two categories to evaluate whether prompts correspond to single API options (one-to-one) or multiple API options (one-to-many), with the performance results shown in Table 7. Surprisingly, the one-to-one scenario performs significantly worse than the one-to-many scenario, with average performance gaps of 21.90\% and 16.14\% for the 7B and 13B models, respectively. Additionally, the one-to-one scenario exhibits a higher hallucination rate. Detailed performance comparisons for one-to-one and one-to-multi scenarios across different input modality types can be found in the Appendix. Our analysis suggests that the number of one-to-one pairs is relatively small, and those one-to-many instances are split into multiple one-to-one instances during training, it is more difficult for the system to predict one-to-one instances correctly. Also, one-to-one pairs have a greater risk of overfitting, where the model becomes overly tuned to specific pairs, hindering its ability to generalize to slightly different queries or contexts during inference. The higher rate of hallucinations also verifies the speculation.

\textbf{Individual modality}. We further examine the model's understanding of instructions with each modality, as shown in Table \ref{tab:suppl_one2mul}. Among them, audio achieves outstanding performance, with 93.29\% accuracy on both Llama2-7B and Vicuna-7B, highlighting its strength in instruction comprehension. In contrast, pure text performance is less satisfactory, with limitations in accuracy and higher hallucination rates. This underperformance may stem from the complexity of text-based tasks, as large models handle multiple tasks or fine-grained sub-tasks. For instance, in multilingual translation, models must evaluate whether the specific case aligns with the API's capabilities, complicating decision-making and affecting performance. Video performs the worst, likely due to limited data and the complexity of video tasks. The model may overemphasize fine-grained details as training progresses, impacting its generalization ability.
\subsection{Ablation Study} \label{ablation}
\textbf{Multimodal Encoders.} To validate the effectiveness of introducing multimodal encoders, we compare the tool selection accuracy in Table \ref{tab:ablation}. For methods without using the multimodal encoder, we directly input the text instruction with a placeholder, like``$\langle$Image$\rangle$", and ignore the non-text data. We can observe that with a multimodal encoder, the performance improves significantly, which highlights that relying solely on textual descriptions without incorporating non-textual data is insufficient for achieving optimal results.

\vspace{-3pt}

\begin{table}[htbp]
\begin{footnotesize}
\centering
\renewcommand{\arraystretch}{1.0}
\resizebox{1.0\columnwidth}{!}{
    \begin{tabular}{ccccc}
    \hline
      & Vicuna-7B & Llama-7B & Llama2-7B & Llama2-Chat-7B \\
    \hline
     w/ multimodal encoder & 83.15& 84.48 & 80.68 & 83.60  \\
    \hline
     w/o multimodal encoder &69.29 & 68.96  & 70.63 & 69.29  \\
    \hline
    & Vicuna-13B & Llama-13B & Llama2-13B & Llama2-Chat-13B \\
    \hline
     w/ multimodal encoder & 88.19 & 87.86  & 81.81 & 87.78 \\
    \hline
     w/o multimodal encoder &69.21 & 69.13 & 68.59 & 69.13  \\
    \hline
    \end{tabular}}%
    \caption{The ablation study on the multimodal encoder reveals a significant performance improvement when the multimodal encoder is utilized. In contrast, without the multimodal encoder, the performance stagnates within a limited range.}
    \vspace{-15pt}
\label{tab:ablation}
\end{footnotesize}
\end{table}%
\section{Conclusion}
\vspace{-3pt}
In this work, we introduce how to empower the tool-augmented LLMs with perceiving visual- or audio-grounded instruction information. We collect a benchmark, ToolMMBench, from the HuggingFace Platform, which covers 932 high-quality machine learning-related APIs. Finetuning this dataset on various popular LLMs has yielded promising results, demonstrating that MLLM-Tool can accurately recommend appropriate tools for multi-modal instructions.

\boldparagraph{Limitations}.
Limited by the computation resources, MLLM-Tool now only supports single-turn conversations and independent multi-turn conversations. Since we directly perceive the content across multi-modal, enabling multi-turn conversations involves encoding the input and output of each modality for every turn, which brings more challenges, including insufficient window size to process and low overall encoding efficiency. Also, although we currently focus on API-type prediction, as predicting the correct API remains a complex task and current performance in this area is suboptimal, the collected dataset provides additional valuable information for the community to utilize.

\boldparagraph{Acknowledgements}.
The authors would thank Qianyu Chen, Haonan Mai and Jindi Guo for constructive feedback and all contributors who participated in our study and assisted in data collection. This work was supported by NSFC $\#$61932020, $\#$62172279, Program of Shanghai Academic Research Leader, and “Shuguang Program” supported by Shanghai Education Development Foundation and Shanghai Municipal Education Commission.

\clearpage
\setcounter{page}{1}
\maketitlesupplementary
In this supplementary material, we list i) dataset details, including selection criteria, searching for multimodal data, prompt templates, data distribution, and crowdsourcing details; ii) experiment setup, data preprocessing; iii) additional experimental results; iv) visualization of the model card, instruction-answer pairs and the actual cases of when using MLLM-Tool; v) our proposed categorization system.

\section{Dataset Details}
In this section, we elaborate on the dataset construction details and the training and testing data distribution. Precisely, Figure \ref{sec:visual_data} visualizes the entire dataset construction process, while \cref{sec:filter} - \cref{sec:prompt} detail each step, including API selection criteria, API functional boundaries criteria, and prompt construction. \cref{sec:visual_distribution} visualizes some data distributions of the training set and testing set. \cref{sec:crowdsourcing} describes the process of employing annotators and instructing them on some guidelines, and in \cref{sec:collection}, we explain the details when searching for content-matched non-text data. We emphasize that the collection and processing of this dataset involved over 2000 person-hours, ensuring its reliability for advanced computer vision research and applications by meticulous attention to detail.

\begin{figure}[htbp]
	\centering
	\includegraphics[width=0.9\linewidth]{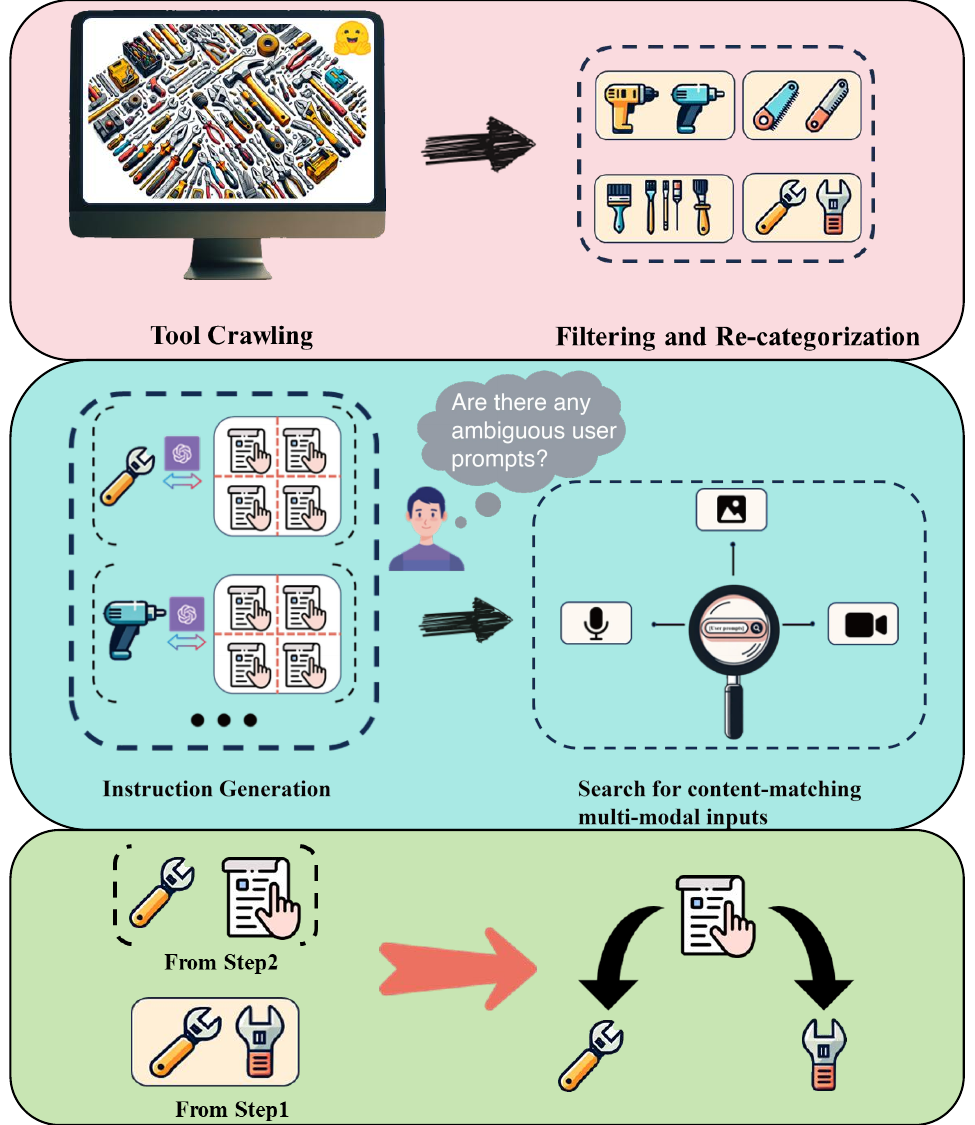}
	\caption{The process of dataset construction. It mainly consists of three stages. The first step is collection and preprocessing, including crawling the original API from HuggingFace and filtering and re-categorizing the API according to tasks. The second step is the generation of instruction-answer pairs, including using GPT-4 to generate instructions based on API functions, determining whether there is text ambiguity in the user prompts, and searching for content-matching input for multi-modal. The third step is constructing a potential one-to-many instruction-answer pair based on the relationship between API functions. }
	\label{fig:dataset1}
\end{figure}

\subsection{Visualization of Dataset Construction Process} \label{sec:visual_data}
\begin{figure}[ht!]
	\centering
	\includegraphics[width=0.6\linewidth]{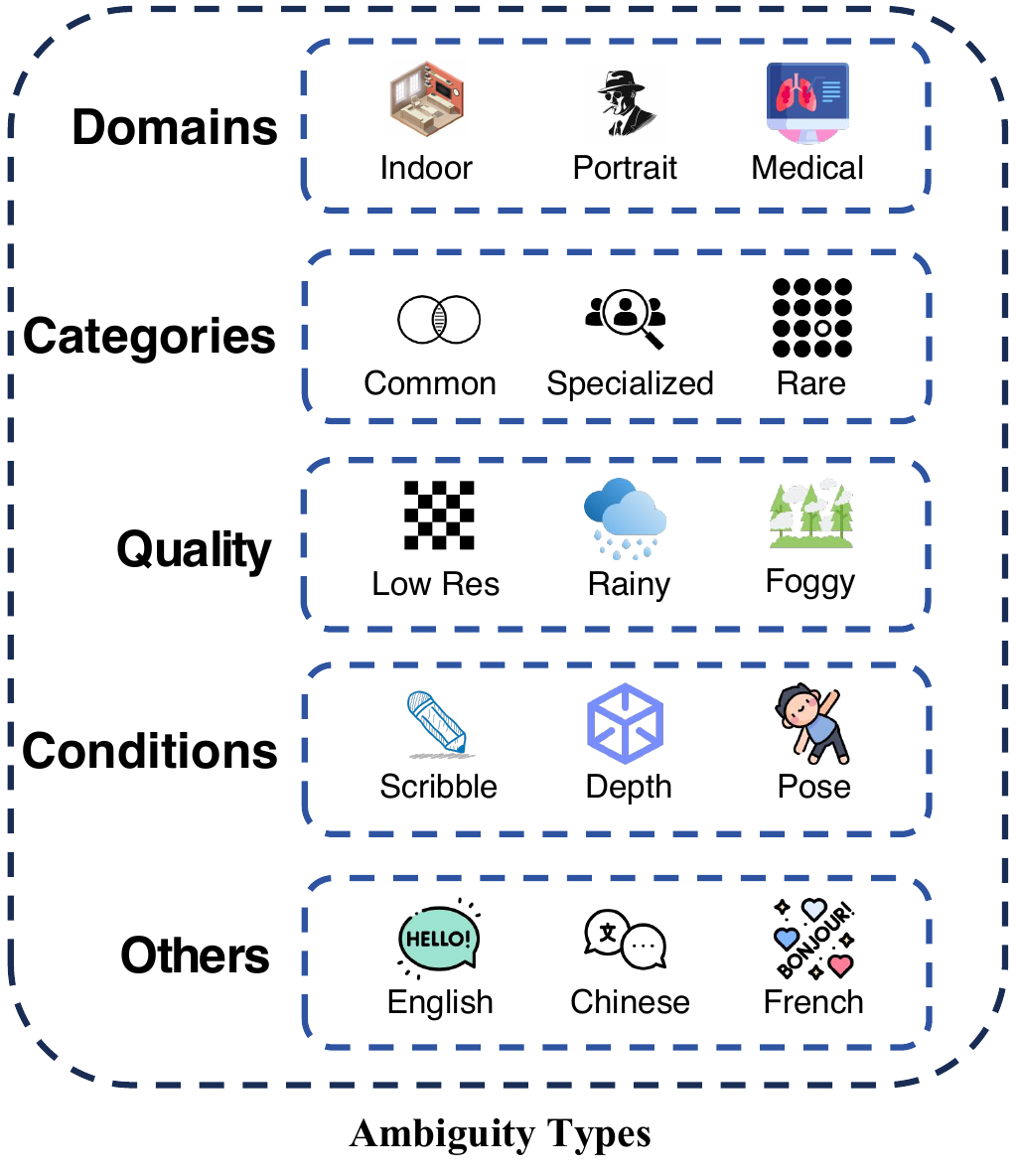}
	\caption{Five types of ambiguity cases need to be checked during the instruction-answer pair generation step.}
	\label{fig:dataset}
\end{figure}

We show the whole process of collecting our dataset in Figure \ref{fig:dataset1}. The difficulty in collecting the benchmark is reflected in three aspects. (1) Task Classification: We restructure a hierarchical task classification system. This advancement rectifies the overly coarse-grained categorizations observed in the HuggingFace platform, ensuring each API has a specific and unambiguous task label. (2) Annotation of Multimodal Input Instructions: Should an instruction reveal cues from modalities other than text, a precise coherence between the divulged information in the instruction and the content of the non-textual modality is paramount. In cases where no such cues are disclosed, we deem it essential to investigate five potential cases caused by textual ambiguity. Moreover, we show the visualization of five ambiguity types in Figure \ref{fig:dataset}. Such scrutiny lays the groundwork for subsequent experiments aimed at validating whether information from other modalities aids large language models in making appropriate tool selections; (3) Instruction Matching: To identify all APIs meeting an instruction's functional requirements, we must categorize APIs based on identical and similar functionalities respectively during collection. When pairing instructions for groups of APIs with analogous functions, it becomes crucial to discern whether a provided instruction strictly satisfies each API's functional description. This determination will influence the number of APIs that can aptly respond as the correct answer to a given instruction.

\subsection{Filtering Rules} \label{sec:filter}
After crawling APIs from HuggingFace, we implement stringent selection criteria for APIs due to the varying quality of these models. The specific filtering rules for this selection process are outlined as follows:

\textbf{Low-Quality Model Card}. Many APIs either lack model cards or contain overly simplistic ones, providing little helpful information. Additionally, some APIs have ceased maintenance or have been integrated into other APIs. Even multiple APIs share a single model card. These APIs pose challenges in organizing API documentation and require manual consultation of additional sources like papers and GitHub. While enriching the number of APIs, this approach introduces inaccuracies. Finally, we opted to remove such APIs from our dataset uniformly.

\textbf{NSFW Content Risk}. For ethical and safety considerations, we remove APIs, particularly some text-to-image types, which pose risks of generating NSFW (Not Safe For Work) content. Such content is deemed inappropriate for all age groups and could compromise the dataset's applicability in diverse settings.

\textbf{License Restrictions}. We exclude models with restrictive licensing terms. An example is the Llama2 model, which requires a formal application and approval process to download and use its model weights.

\textbf{Identical APIs}. Numerous models on HuggingFace share identical structures and datasets, merely differing in implementation by various contributors. To avoid redundancy, we selectively retain only the most downloaded version of each such model, ensuring the dataset is more streamlined. For APIs with different model architectures but trained on the same dataset and having the same functions, as we mentioned in the main text, we select the top five APIs based on download statistics.

\subsection{API Function Boundaries} \label{sec:boundary}
\textbf{Multi-task APIs}. Our dataset includes multi-task APIs like various large language models (Bloom ~\cite{workshop2022bloom}, Baichuan2 ~\cite{yang2023baichuan}, Falcon ~\cite{penedo2023refinedweb}), capable of numerous NLP tasks, some of which even fall outside HuggingFace's classification system. For simplicity, we focus specifically on their text generation capabilities, aligning with HuggingFace's functional categorization of such APIs. Additionally, we modify the descriptions of these APIs in our dataset to ensure a match between the functionalities reflected by the provided queries and those described in the API descriptions.

\textbf{API Function Differentiation}. After identifying the specific subdomain of each API, we detail our approach for discerning whether APIs within the same subdomain possess identical or similar functions. For plain text input APIs, we can assess APIs based on the language and context of the datasets they utilize. For example, in sentiment analysis tasks, we discern various contexts by relying on the distinct domains of data sources, such as financial, legal, social media, and reviews, enabling us to classify nuanced functional differences among similar-function APIs. Similarly, we refine API function categorization for multimodal inputs, especially images, by analyzing dataset contexts. For instance, in Visual Question Answering tasks, datasets feature a range of images such as Diagrams, Charts, Documents, Infographics, and other general images, enabling nuanced API segmentation. The above examples provide merely a subset of our API classification rules. Due to the complex and case-by-case nature of the analysis, the complete set of rules is not listed here.

\begin{figure*}[htbp]
\vspace{-6pt}
  \centering
      \hfill 
    \begin{subfigure}{0.35\textwidth}
      \centering   
      \includegraphics[width=\textwidth]{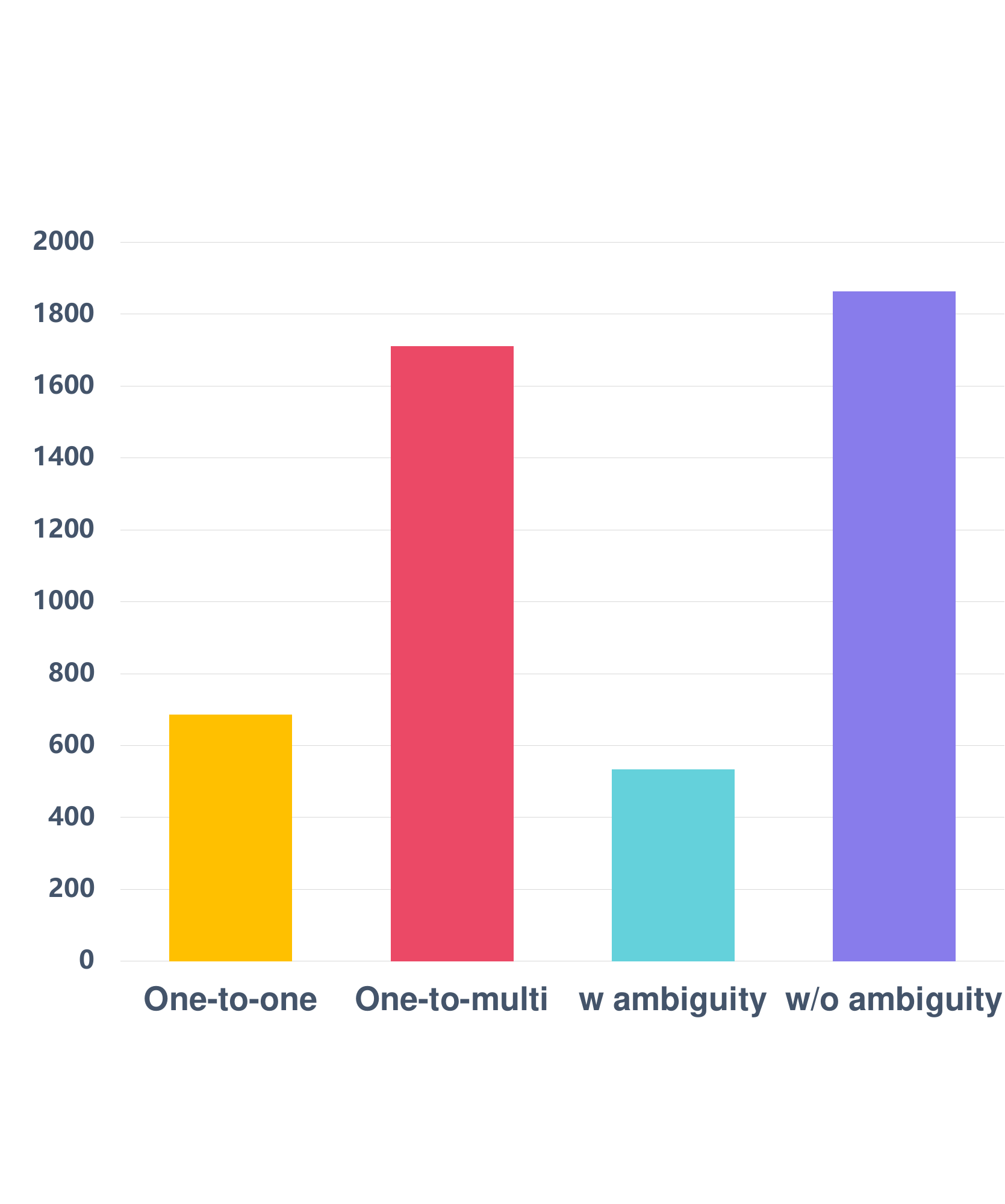}
      \caption{The distribution of sub-testing sets divided by API option number and ambiguity types respectively.}
      \label{fig:subset}
    \end{subfigure}        
    \hfill 
    \begin{subfigure}{0.3\textwidth}
      \centering   
      \includegraphics[width=\textwidth]{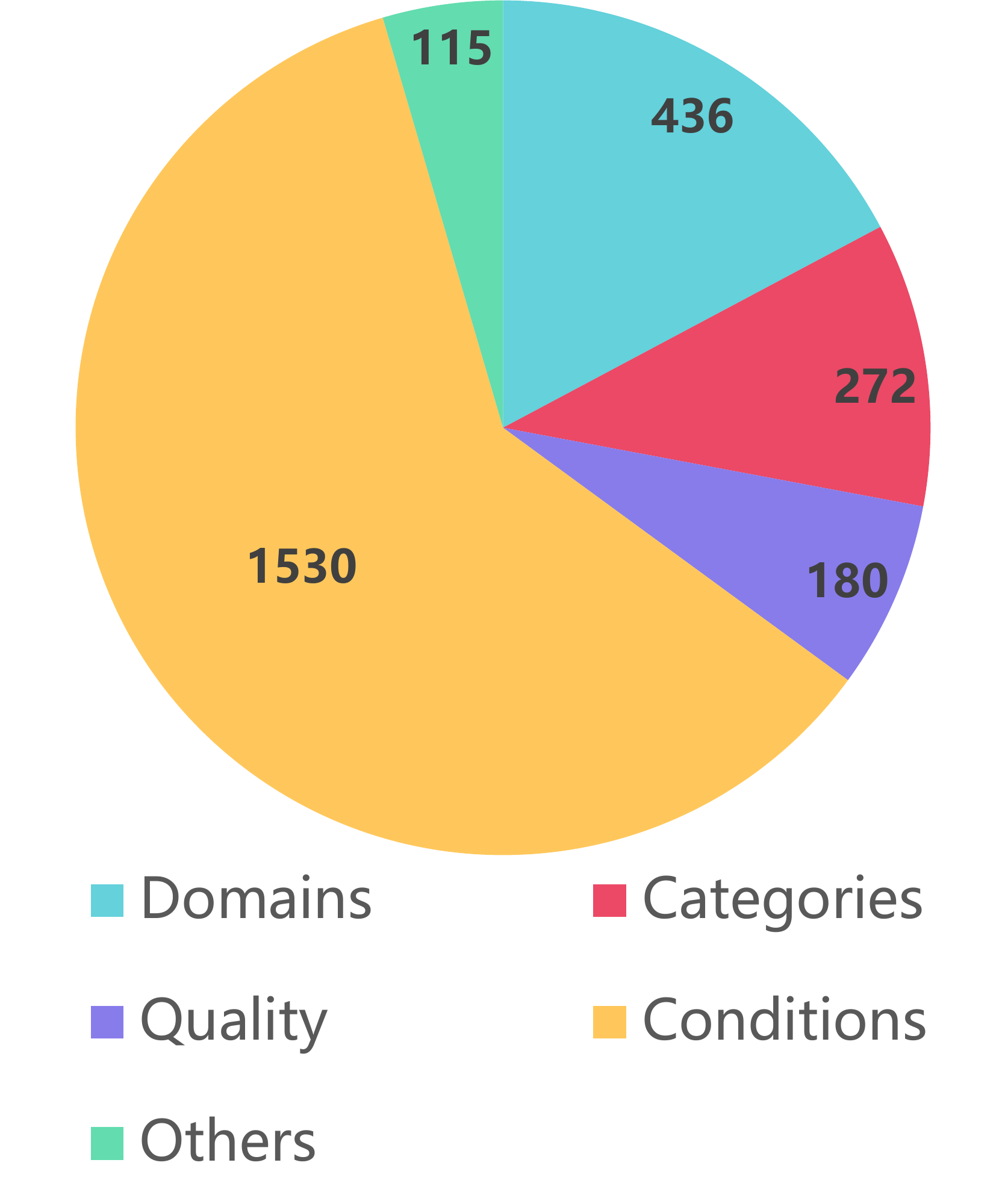}
      \caption{The distribution of each ambiguity type in the whole dataset.}
      \label{fig:ambiguity_all}
    \end{subfigure}        
    \hfill
    \begin{subfigure}{0.3\textwidth}
      \centering   
      \includegraphics[width=\textwidth]{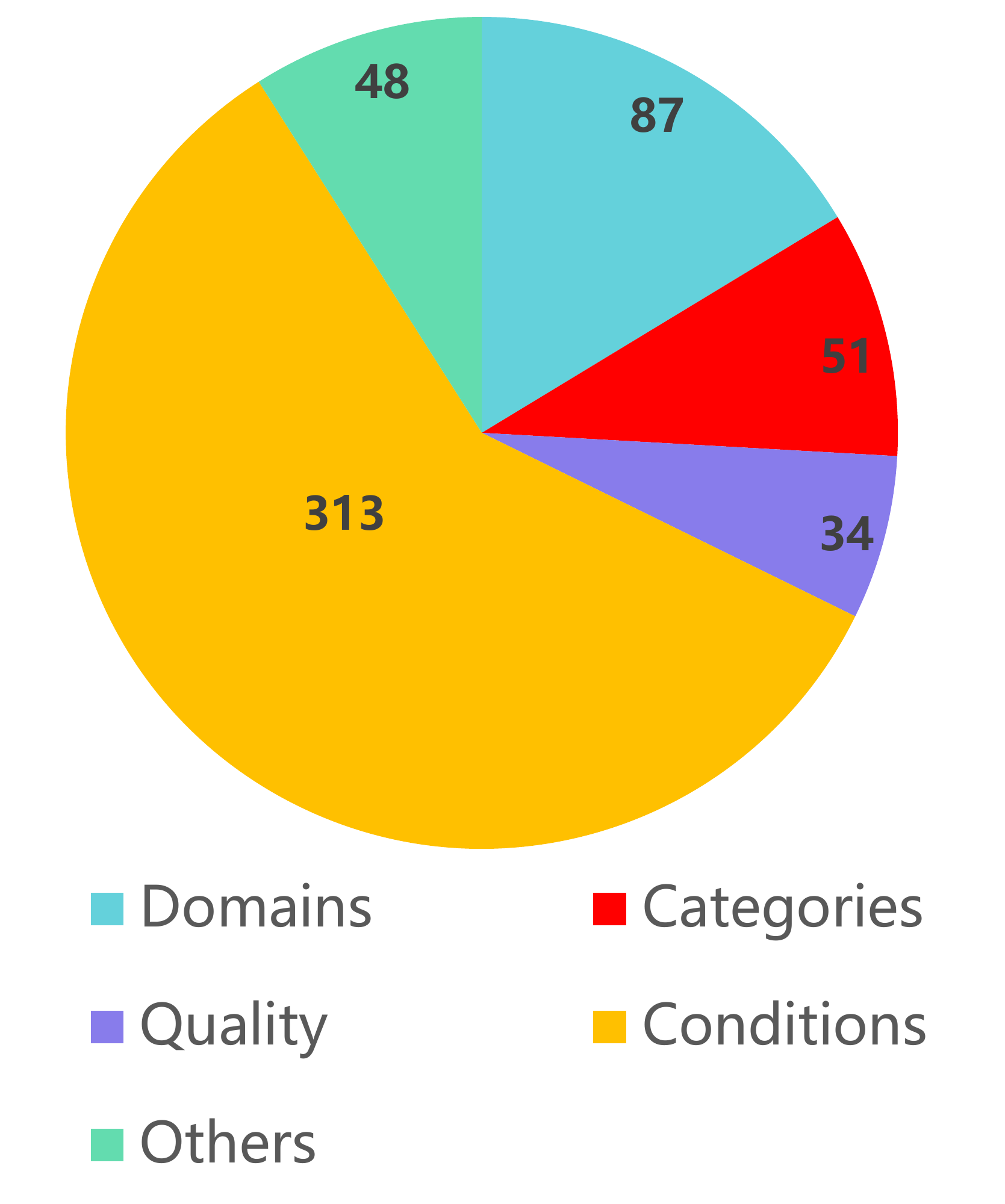}
      \caption{The distribution of each ambiguity type in the testing set.}
      \label{fig:ambiguity_test}
    \end{subfigure}        
    \hfill
\vspace{-1pt}
\caption{
\textbf{Dataset statistic visualization.} 
}
\label{fig:dataset_config}
\vspace{-15pt}
\end{figure*}

\textbf{One-to-many Situation Identification}. 
Our approach, driven by GPT-4, originates from each API's functional description and generates corresponding instructions that adhere strictly to the API's functional boundaries. Hence, consideration is primarily given to other APIs with the same function in one-to-many scenarios.

\subsection{Prompt Construction} \label{sec:prompt}
Below, we list the detailed prompts used for generating instructions. The prompts fed to GPT-4 are composed of (1)the API call's relevant information, including its name, a concise functional description, and training datasets if mentioned in the model card, (2)some sample instructions, and (3)additional requirements. The sample instructions are leveraged to foster format standardization and enhance the quality of GPT-4's outputs. So, we invite some expert annotators to craft high-quality instructions in advance and randomly choose two each time as exemplary inputs to GPT-4. We design two sets of prompt templates in our dataset, mindful of prompt ambiguity cases. For APIs with multimodal inputs susceptible to prompt ambiguity, we strive to balance the final selection of ten instructions, maintaining a 1:1 ratio between unambiguous and ambiguous instructions; the corresponding prompt template is shown in Figure \ref{fig:data_exp} and Figure \ref{fig:data_exp_amb} respectively. We emphasize that query ambiguity pairs are pairs with identical textual inputs but varied multimodal inputs.

\subsection{Visualization of Data Distributions} \label{sec:visual_distribution}

We tally the instances of each ambiguity type in Figure \ref{fig:ambiguity_all}, amassing a total of 2533 ambiguity pairs. Predominantly, these are concentrated within conditional cases involving image inputs, comprising 1530 pairs, since our dataset covers 11 distinct condition types of ControlNet ~\cite{zhang2023adding}, and their various combinations, achieving a considerable quantity.

Particularly for the testing set, we depict in Figure \ref{fig:subset} and Figure \ref{fig:ambiguity_test} the distribution of API quantities across each testing subset. These subsets are constructed based on three distinct partitioning criteria: ambiguity types, API option number, and modality. It is worth noting that the \textit{without ambiguity} category consists of two parts: cases with purely textual inputs that inherently lack ambiguity (1,452 instances) and cases with non-textual inputs where ambiguity is absent (412 instances).

In our work, we collect 932 APIs and one \textit{"Unknown"} category, featuring 651 APIs with pure text inputs, 191 APIs incorporating image inputs, 80 incorporating audio inputs, and 10 incorporating video inputs. Given the long-tail distribution of API quantities for various tasks on HuggingFace, coupled with our selection criteria, the APIs we ultimately choose exhibit a similar long-tail distribution across both modality and task level. In future work, we aim to enrich our collection with more APIs, particularly those with video inputs, and optimize the distribution across different tasks for more uniformity.

\subsection{Crowdsourcing Details}
\label{sec:crowdsourcing}
In this study, we employ eight annotators with NLP or CV backgrounds, with an estimated hourly 7.10\$ compensation and an up to 20\% bonus for those with high annotation quality. We provide a comprehensive training document and record a video to guide them through the process of annotation. We also instruct them to avoid using private information or license-risky data from unknown or sensitive sources. All annotators are presented with a consent form and must agree to join the project. We set up a cross-checking mechanism to correct errors.
Moreover, our team members conduct manual checking and optimization. If the data is collected from a website, we restore the website links and upload the metadata when releasing the dataset. We claim we only use the data for academic research instead of commercial usage. 

\subsection{Collection of text-matched multimodal input}
\label{sec:collection}
Our non-text data sources are from either the dataset API used or Google Images. We prioritize collecting from open-source datasets by manually checking that the data content is consistent with the instructions generated by GPT. For the latter, we can retrieve the data by keywords in the Google database, and for these data, we pay more attention to its license to ensure its legality.

\begin{table}[!htb]
    \begin{center}
        \centering
        \resizebox{\linewidth}{!}{
        \begin{tabular}{c|c|c}
            Configuration & 7B-model & 13B-model \\
            \midrule
            Optimizer & AdamW & AdamW\\
            Optimizer Momentum & $\beta_1=0.9, \beta_2=0.95$ & $\beta_1=0.9, \beta_2=0.95$\\
            Peak learning rate & 5e-4 & 5e-4\\
            Weight decay & 0.001 & 0.001\\
            Warmup steps & 10 & 10 \\
            Batch size & 64 & 32\\
            Micro-batch size & 4 & 1\\
            Gradient accumulation steps & 4 & 8\\
            Maximum target length & 512 & 512\\
            ImageBind Checkpoint & ImageBind-Huge & ImageBind-Huge \\
            \hline
            LoRA attention dimension (r) & 32 & 32\\
            LoRA scaling alpha ($\alpha$) & 32 & 32\\
            LoRA drop out & 0.1 & 0.1
        \end{tabular}}
    \end{center}
    \caption{7B and 13B model training configuration.}
    \label{tab:setup}
    \end{table}

\section{Experiment Setup and Data Processing}
We list the setup, including some hyperparameters used to train our model, in Table \ref{tab:setup}. We adopt ImageBind's data processing approach for the multimodal inputs. We evenly split the video into five clips and randomly sampled two frames from each clip to cover the entire length, similar to other Vision Transformer (ViT)-based works with video input. For audio, we sample each input audio at 16KHz. Subsequently, we capture a log mel spectrogram featuring 128 frequency bins. Then, the spectrogram can be seen as a 2D image. In this way, we can use ViT for these three multimodal encoders. A modality-specific linear projection head is added to each modality's encoder to align the feature into a uniform 1024-sized dimension.

\section{Additional Experimental Results}

\subsection{Results of Multiple Options Categorized by Input Modality}
Regarding the condition of whether multiple API options exist, we further refine the testing subsets—originally divided based on the presence or absence of multiple API options—by introducing an additional criterion based on input modality to gain deeper insights into the issue. We select two versions of the Vicuna model with different parameter sizes, and the corresponding performance is presented in Table \ref{tab:suppl_onetomany}. From the results, we observe that for the case where multiple API options exist, the performance is consistently better than the case with only one API option, regardless of the input modality. This validates the argument presented in the main text. On one hand, the presence of multiple API options reduces the problem's complexity and increases the likelihood of selecting the correct API. On the other hand, during training, the data involving multiple API options benefits from being split into multiple one-to-one training strategies, further improving performance.

\begin{table}[htbp]
\renewcommand{\arraystretch}{0.7}
\resizebox{1.0\columnwidth}{!}{
    \begin{tabular}{cccc}
    \hline
    Model  & API Options & Input Modality & Acc  \\
    \hline

    Vicuna-7B  & one-to-one  & Video  & 55.56  \\
    Vicuna-7B  & one-to-many  & Video  & 83.33  \\
    Vicuna-7B  & one-to-one  & Audio  & 78.57  \\
    Vicuna-7B  & one-to-many  & Audio  & 96.32  \\
    Vicuna-7B  & one-to-one  & Image  & 75.82  \\
    Vicuna-7B  & one-to-many  & Image  & 96.30  \\
    Vicuna-7B  & one-to-one  & Text  & 58.24  \\
    Vicuna-7B  & one-to-many  & Text  & 86.43  \\
    \hline

    Vicuna-13B  & one-to-one  & Video & 44.44\\
    Vicuna-13B  & one-to-many  & Video & 91.67\\
    Vicuna-13B  & one-to-one  & Audio & 75.00\\
    Vicuna-13B  & one-to-many  & Audio & 95.59\\
    Vicuna-13B  & one-to-one  & Image & 86.81\\
    Vicuna-13B  & one-to-many  & Image & 96.10\\
    Vicuna-13B  & one-to-one  & Text & 70.21\\
    Vicuna-13B  & one-to-many  & Text & 90.24\\
    \hline
    \end{tabular}}%
    \centering
  \caption{The performance of the testing subsets based on the joint criterion of the presence of multiple API options and the input modality.}
  \label{tab:suppl_onetomany}%
\end{table}%

\begin{figure*}
\begin{tcolorbox}
    \textbf{\#\#\# User}: You are an NLP task expert. Given an API, you need to generate 20 different user queries that utilize the API function, adhering to the following input and output format for each query: \\
    Input: \\
    API Name: This is the name of the API Function. \\
    Description: This is a detailed description of the model. \\
    Prohibit Words: These are words that are prohibited from appearing in the output.\\

    Output: \\
    Query1: This is an instruction that can utilize the API function.\\
    Query2: This is an instruction that can utilize the API function.\\
    Query3: This is an instruction that can utilize the API function.\\
    Query4: This is an instruction that can utilize the API function.\\
    Query5: This is an instruction that can utilize the API function.\\
    \dots \\
    Query20: This is an instruction that can utilize the API function.\\

    Below are some examples: \\
    Example 1:\\
    Input:\\
    API Name: microsoft/codereviewer \\
    Description: CodeReviewer is a model pre-trained with code change and code review data to support code review tasks.\\
    Prohibit Words: "API, tools, model"\\
    
    Output:\\
    Query1: I've made some changes to my Python code. Can you review this snippet for me? "def addNums(a, b): return a + b"\\
    Query2: I'm not very confident in my Java coding skills. Could you check this piece for potential issues? "public int divide(int x, int y) \{ return x / y; \}"\\
    Query3: I'm trying to refactor this C\# method. Does it look okay? "public void PrintName(string name) \{ Console.WriteLine(name); \}"\\
    Query4: I wrote this SQL query, and I'm unsure if it's optimized. Can you review it? "SELECT * FROM users WHERE age $\geq$ 25;"\\
    Query5: Here's a piece of JavaScript function I came up with. Any suggestions for improvement? "function greet(name) \{ return 'Hello ' + name; \}"\\
    Query6: I've been learning Ruby recently. Would appreciate your thoughts on this code segment: "def multiply(x, y) x * y end"\\
    Query7: I'm not sure if this CSS is correct. Can you review it? "h1 \{ color: blue; font-size: 30px; \}"\\
    Query8: I just started with Swift. Here's a function I wrote. Could you review it for best practices? "func displayMessage(message: String) \{ print(message) \}"\\
    Query9: My colleague wrote this PHP function, but it seems off. Can you review it for me? "function subtract($x, $y) \{ return $x - $y; \}"\\
    \dots \\
    Query20: I'm learning Kotlin and wrote this simple method. Mind taking a look? "fun sum(a: Int, b: Int): Int = a + b" \\

    \textit{To be continued}

\end{tcolorbox}
\end{figure*}

\begin{figure*}
\begin{tcolorbox}
\textit{Continued}\\
    Example 2:\\
    Input:\\
    API Name: Salesforce/blip-vqa-base \\
    Description: This model is a base-sized ViT-based BLIP trained on the visual question answering task VQA2.0 dataset. The model's input is an image and a text question, and the model's output is a text answer.\\
    Prohibit Words: "API, tools, model, VQA2.0"\\
    Output:\\
    Query1: I recently visited a new city and took some photographs. Based on this particular image, can you tell me what is the weather conditions like?\\
    Query2: I was at a photo exhibition and took this picture. Could you provide information about was this taken in a Latin American country judging from the dress?\\
    Query3: I was at the park and saw a dog playing. By analyzing this photo, can you tell if the frisbee it is using is hard or soft?\\
    Query4: My cousin went to a new food joint and ordered their specialty hot dog. By looking at this picture, can you determine if the hot dog is larger than a normal one?\\
    Query5: I attended a family gathering and took a group photo. By analyzing the photograph, how many kids can you see in the picture?\\
    Query6: Looking at this captivating portrait, I noticed an object in the individual's hand. It seems to be a sweet treat. What is this person holding?\\
    Query7: I was at the beach and took a photo of a man skiing. By examining the image, can you tell if the wave is chasing him?\\
    Query8: I visited the local train station and captured an image of a passing train. Can you identify what color the train is from the picture?\\
    Query9: This cityscape presents an old-fashioned vehicle that stands out amidst the modern structures. I'd like to know more about this vehicle. Could you tell me its model and color?"\\
    \dots \\
    Query20: Given the serene backdrop of sailboats and a calm sea, can you tell me how many boats are there? \\

    Note that:\\
    1. When crafting queries, avoid including the API's name; \\
    2. Ensure that the queries are varied and diverse; \\
    3. When processing an input, any words listed in the Prohibited Words must be strictly excluded from the response.\\ 
    
    Now, let’s start. \\
    Input: \\
    API Name: [API Name] \\
    Description: [API Description]\\
    Prohibit Words: [Some words] \\

\end{tcolorbox}
\vspace{-12pt}
\caption{\textbf{Example of the unambiguous prompt template}:  This template applies to pure text inputs and a part of multimodal input scenarios where textual inputs reveal multimodal information. In practical usage, the template's masked sections should be populated with the respective API's name, description, and words prohibited in the instructions.}
\label{fig:data_exp}
\end{figure*}

\begin{figure*}
\begin{tcolorbox}
    \textbf{\#\#\# User}: You are an NLP task expert. Given an API, you need to generate 20 different user queries that utilize the API function, adhering to the following input and output format for each query: \\
    Input: \\
    API Name: This is the name of the API Function. \\
    Description: This is a detailed description of the model. \\
    Prohibit Words: These are words that are prohibited from appearing in the output.\\

    Output: \\
    Query1: This is an instruction that can utilize the API function.\\
    Query2: This is an instruction that can utilize the API function.\\
    Query3: This is an instruction that can utilize the API function.\\
    Query4: This is an instruction that can utilize the API function.\\
    Query5: This is an instruction that can utilize the API function.\\
    \dots \\
    Query20: This is an instruction that can utilize the API function.\\

    Below are some examples: \\
    Example 1:\\
    Input:\\
    API Name: timm/resnet101.a1h\_in1k \\
    Description: A 101 layers ResNet-B image classification model trained on ImageNet-1k.\\
    Prohibit Words: "API, tools, model, ImageNet"\\
    
    Output:\\
    Query1: I'd value your input on this image's classification.\\
    Query2: There's an image in my possession, and I'm seeking a label for it.\\
    Query3: I need help in determining the label for this image. Can you help?\\
    Query4: Can you provide a label perspective for this image?\\
    Query5: How would you manage the categorization of this particular visual?\\
    Query6: There's an intriguing image that I've come across, and I'm curious about its category.\\
    Query7: How would you classify this picture?\\
    Query8: I'd be grateful for your view on this image's category.\\
    Query9: I'm searching for a label for this visual. Can you assist?\\
    \dots \\
    Query20: What category seems plausible for this visual, in your view? \\

    Example 2:\\
    Input:\\
    API Name: lllyasviel/control\_v11f1p\_sd15\_depth \\
    Description: This model is intended for control diffusion models by adding extra conditions. Trained with depth estimation, the condition image is an image with depth information, usually represented as a grayscale image, and the output is a new image.\\
    Prohibit Words: "API, tools, model, depth, controlnet"\\

    \textit{To be continued}

\end{tcolorbox}
\end{figure*}

\begin{figure*}
\begin{tcolorbox}
\textit{Continued}\\
    Output:\\
    Query1: Use my image to depict a family picnic in a sunlit park with children playing.\\
    Query2: Can you reimagine my image as a soccer match in progress, with players and fans cheering?\\
    Query3: Create an adventure scene with pirates and treasure islands using my image.\\
    Query4: Can you make my image into a heartwarming scene of a mother duck leading her ducklings through a pond?\\
    Query5: Craft a depiction of a magician performing tricks in front of an amazed audience from my image.\\
    Query6: Could you reshape this photograph into a bustling train station with intricate ironwork and arches?\\
    Query7: Transform this snapshot into a charming cottage nestled among rolling hills and wildflowers.\\
    Query8: I'm curious about how this picture would be reimagined as a bustling harbor with ships, cranes, and cargo containers.\\
    Query9: Use my image to depict a family picnic in a sunlit park with children playing.\\
    \dots \\
    Query20: I'd love to see this image as a gardener carefully pruning roses in a blooming garden. \\

    Note that:\\
    1. When crafting queries, avoid including the API's name; \\
    2. Ensure that the queries are varied and diverse; \\
    3. When processing an input, any words listed in the Prohibited Words must be strictly excluded from the response;\\ 
    4. The queries should not convey or imply multimodal information. \\
    
    Now, let’s start. \\
    Input: \\
    API Name: [API Name] \\
    Description: [API Description]\\
    Prohibit Words: [Some words] \\

\end{tcolorbox}
\caption{\textbf{Example of the ambiguous prompt template}:  This template applies to other multimodal input scenarios where textual inputs do not reveal multimodal information. In practical usage, the template's masked sections should be populated with the respective API's name, description, and words prohibited in the instructions.}
\label{fig:data_exp_amb}
\end{figure*}

\section{Dataset Visualization}

\subsection{Model Card Visualization}
\begin{figure*}[htbp]
	\centering
	\includegraphics[width=\linewidth]{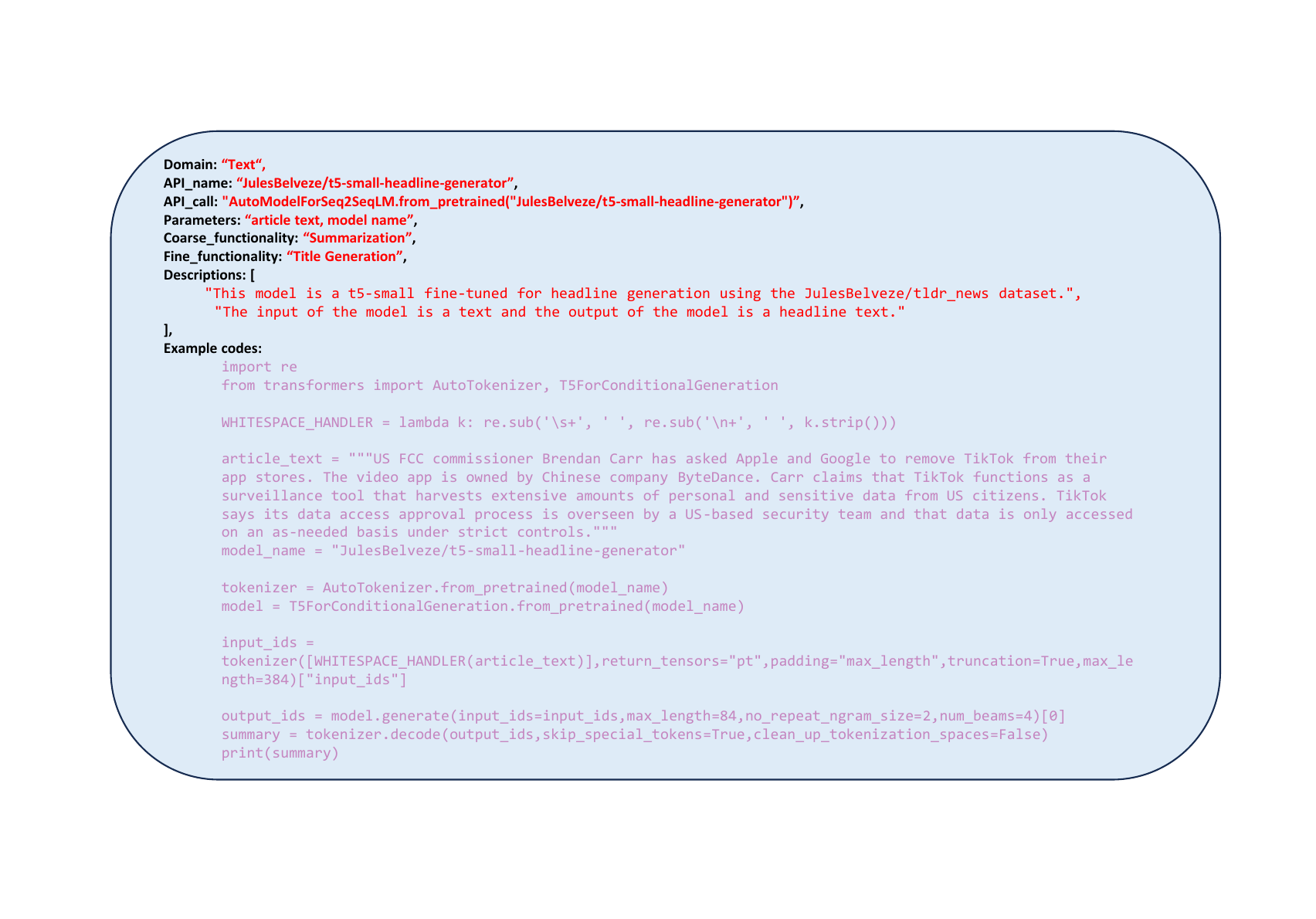}
    \vspace{-60pt}
	\caption{Model Card Visualization. (Sample 1)}
	\label{fig:mc1}
\end{figure*}

\begin{figure*}[htbp]
	\centering
	\includegraphics[width=\linewidth]{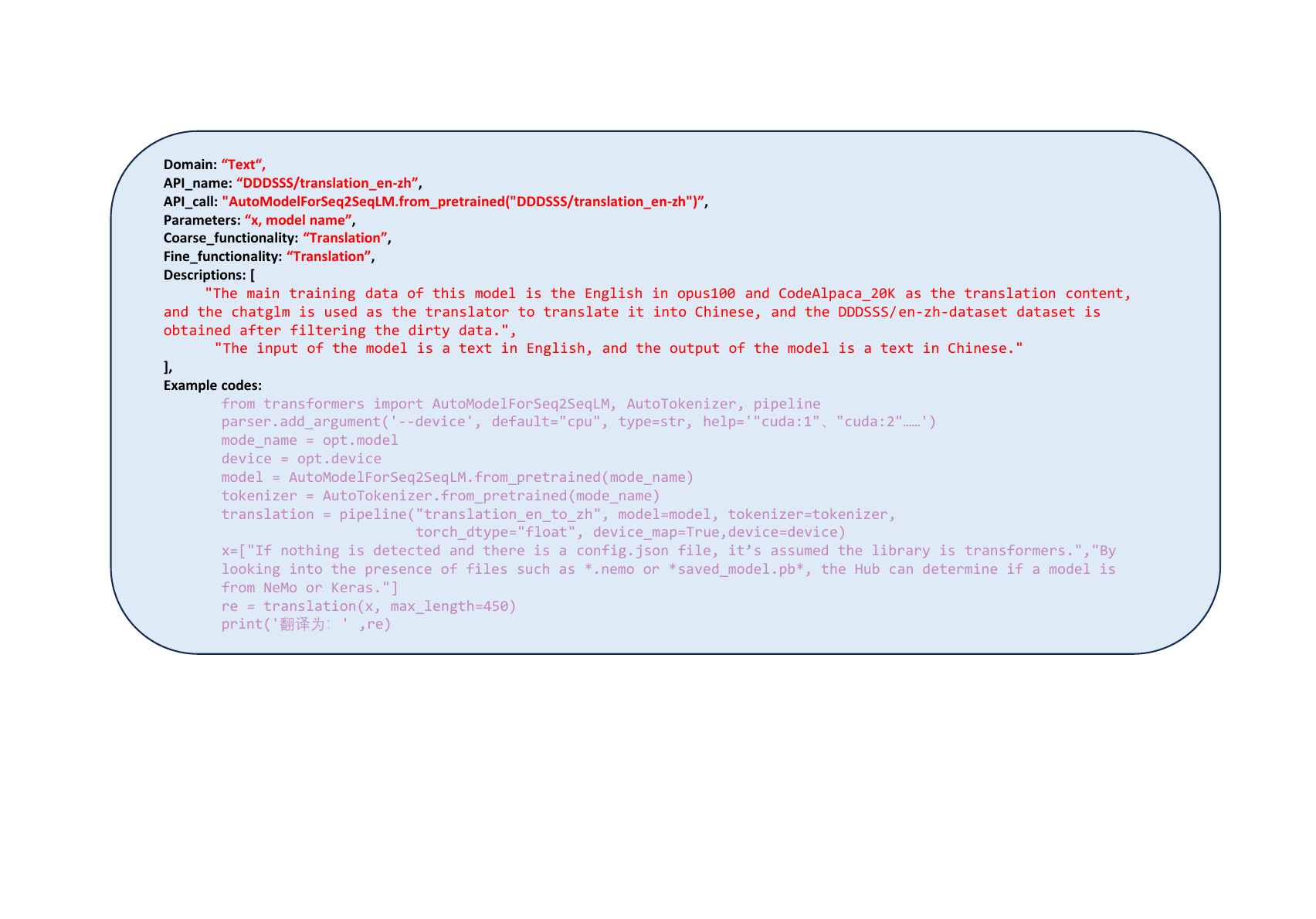}
    \vspace{-120pt}
	\caption{Model Card Visualization. (Sample 2)}
	\label{fig:mc2}
\end{figure*}

\begin{figure*}[htbp]
	\centering
	\includegraphics[width=\linewidth]{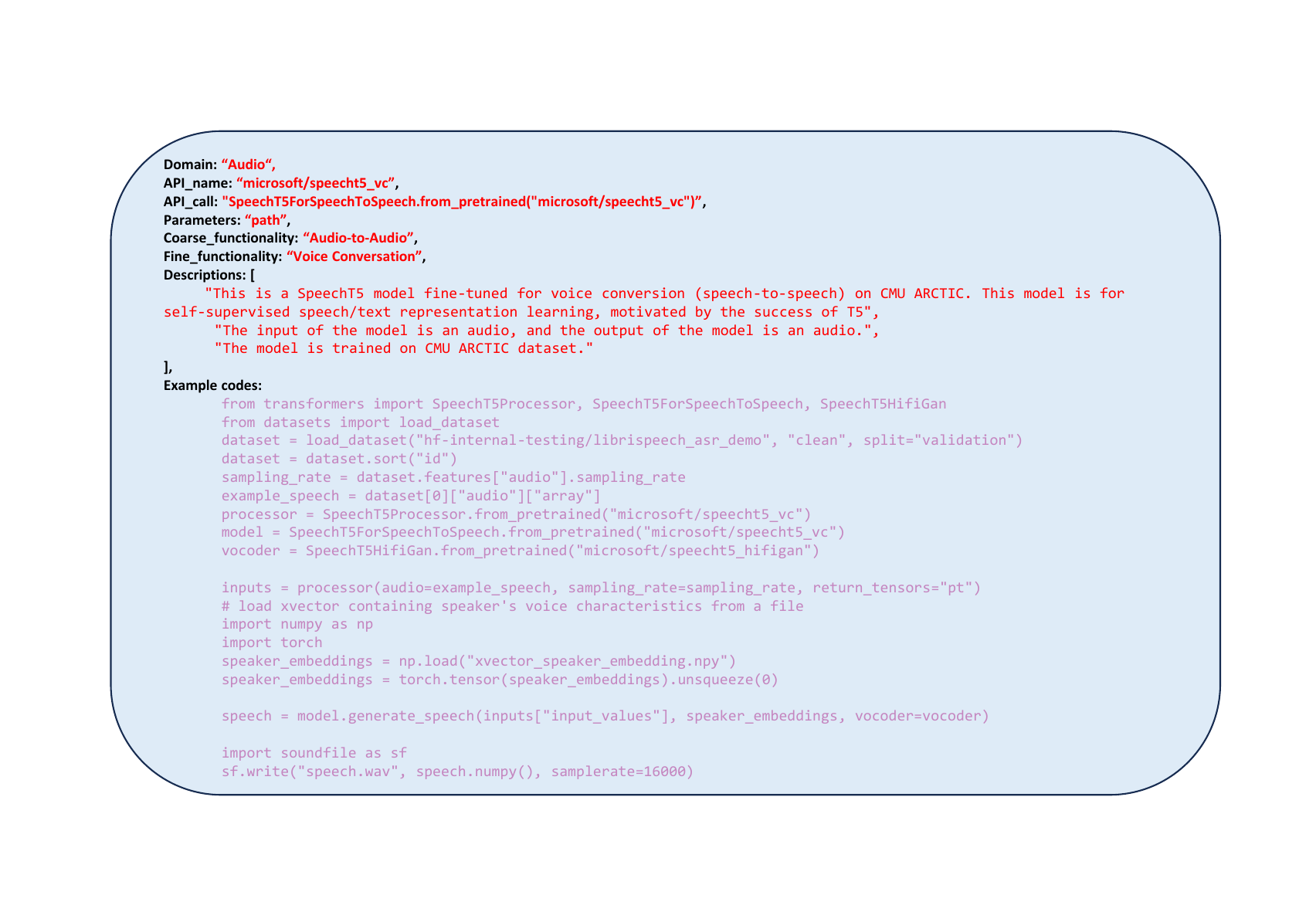}
    \vspace{-65pt}
	\caption{Model Card Visualization. (Sample 3)}
	\label{fig:mc3}
\end{figure*}

\begin{figure*}[htbp]
	\centering
	\includegraphics[width=\linewidth]{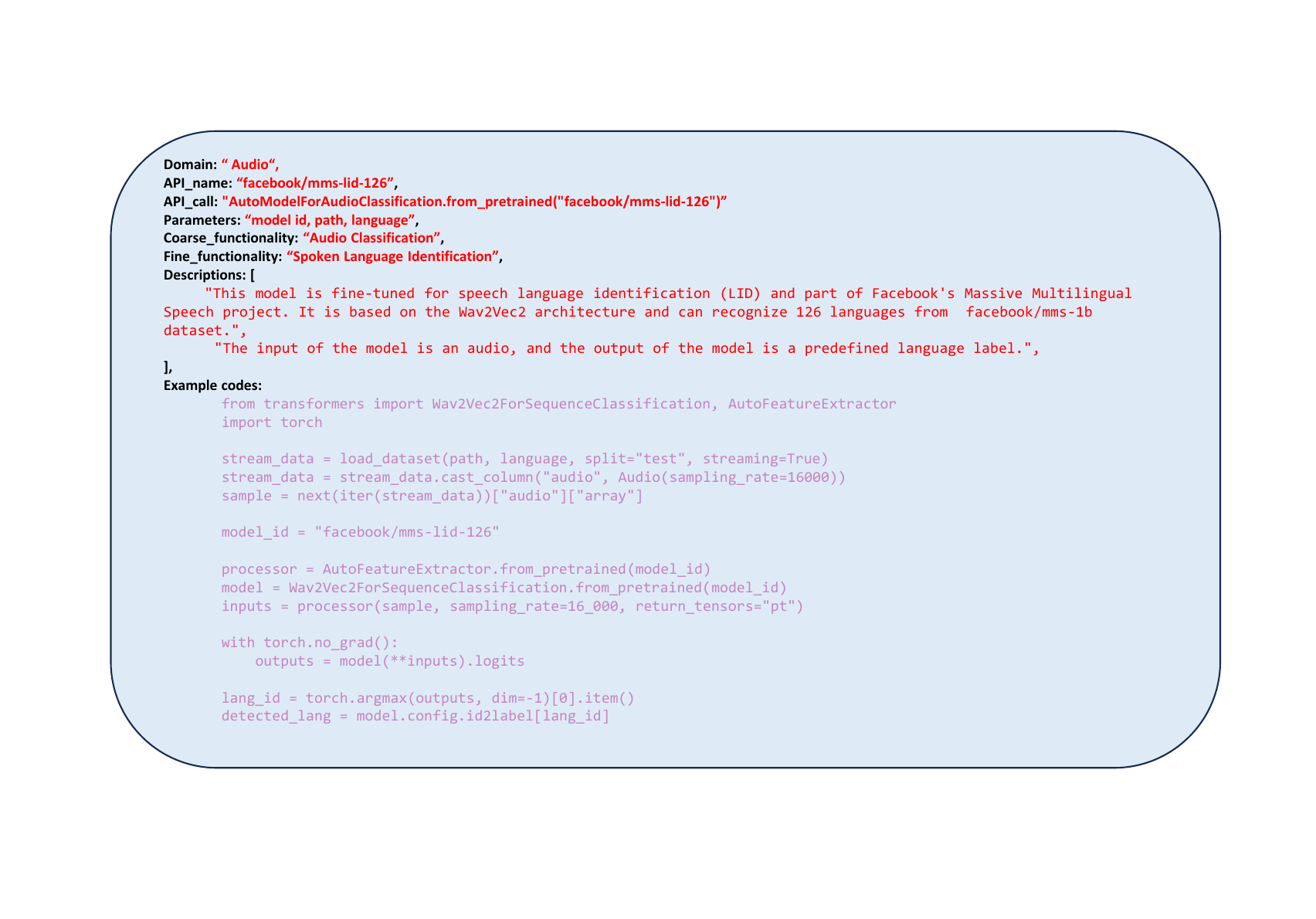}
    \vspace{-70pt}
	\caption{Model Card Visualization. (Sample 4)}
	\label{fig:mc4}
\end{figure*}

\begin{figure*}[htbp]
	\centering
	\includegraphics[width=\linewidth]{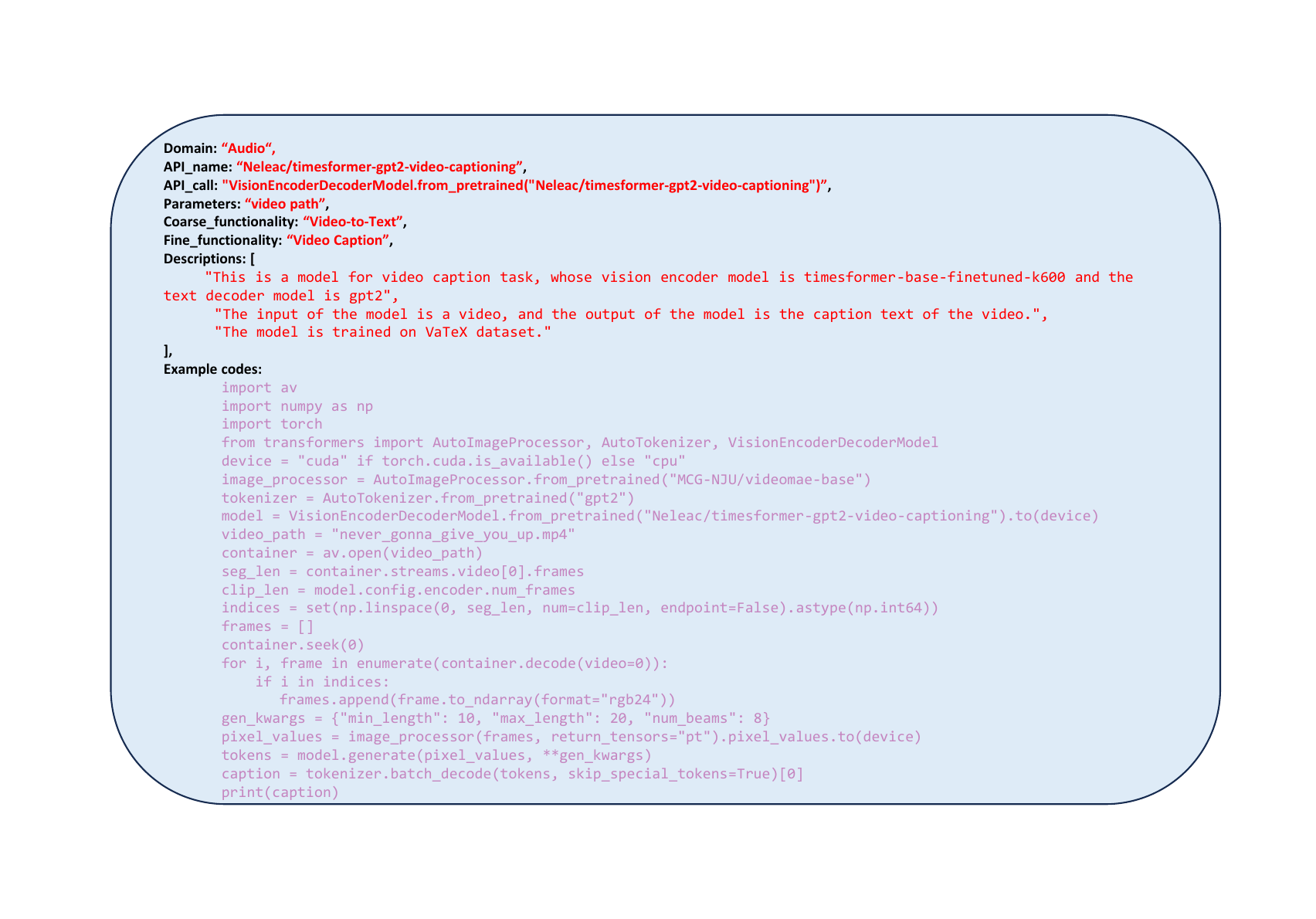}
    \vspace{-60pt}
	\caption{Model Card Visualization. (Sample 5)}
	\label{fig:mc5}
\end{figure*}

\begin{figure*}[htbp]
	\centering
	\includegraphics[width=\linewidth]{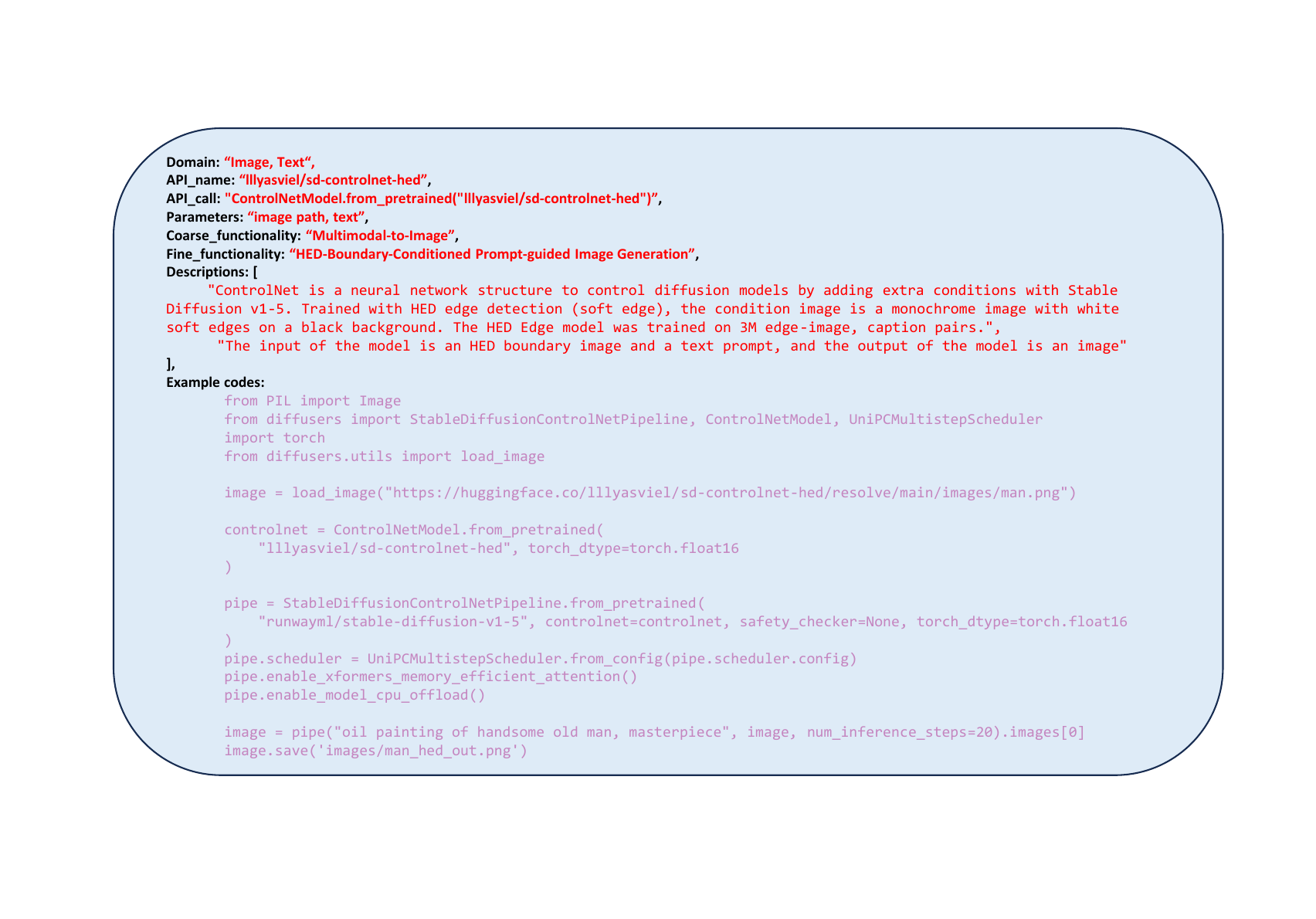}
    \vspace{-70pt}
	\caption{Model Card Visualization. (Sample 6)}
	\label{fig:mc6}
\end{figure*}

We illustrate 6 model card samples in Figure \ref{fig:mc1}-\ref{fig:mc6}, which covers all the information mentioned in the main text. We list some samples from several different modalities and different functions.

\subsection{Instruction-Answer Pairs Visualization}
\begin{figure*}[htbp]
	\centering
	\includegraphics[width=\linewidth]{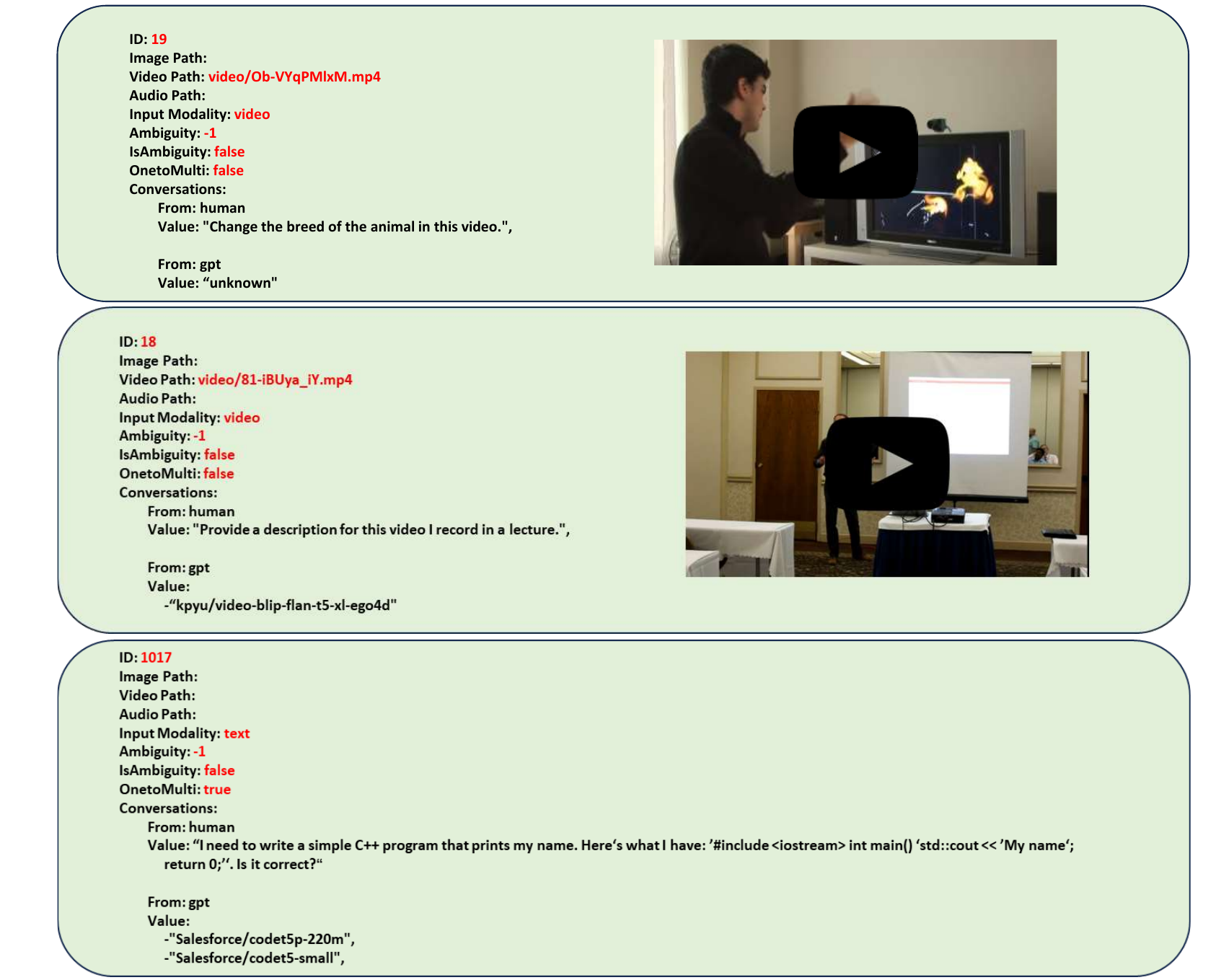}
	\caption{Instruction-Answer Pairs Visualization. (Sample 1-3)}
	\label{fig:iap1}
\end{figure*}

\begin{figure*}[htbp]
	\centering
	\includegraphics[width=\linewidth]{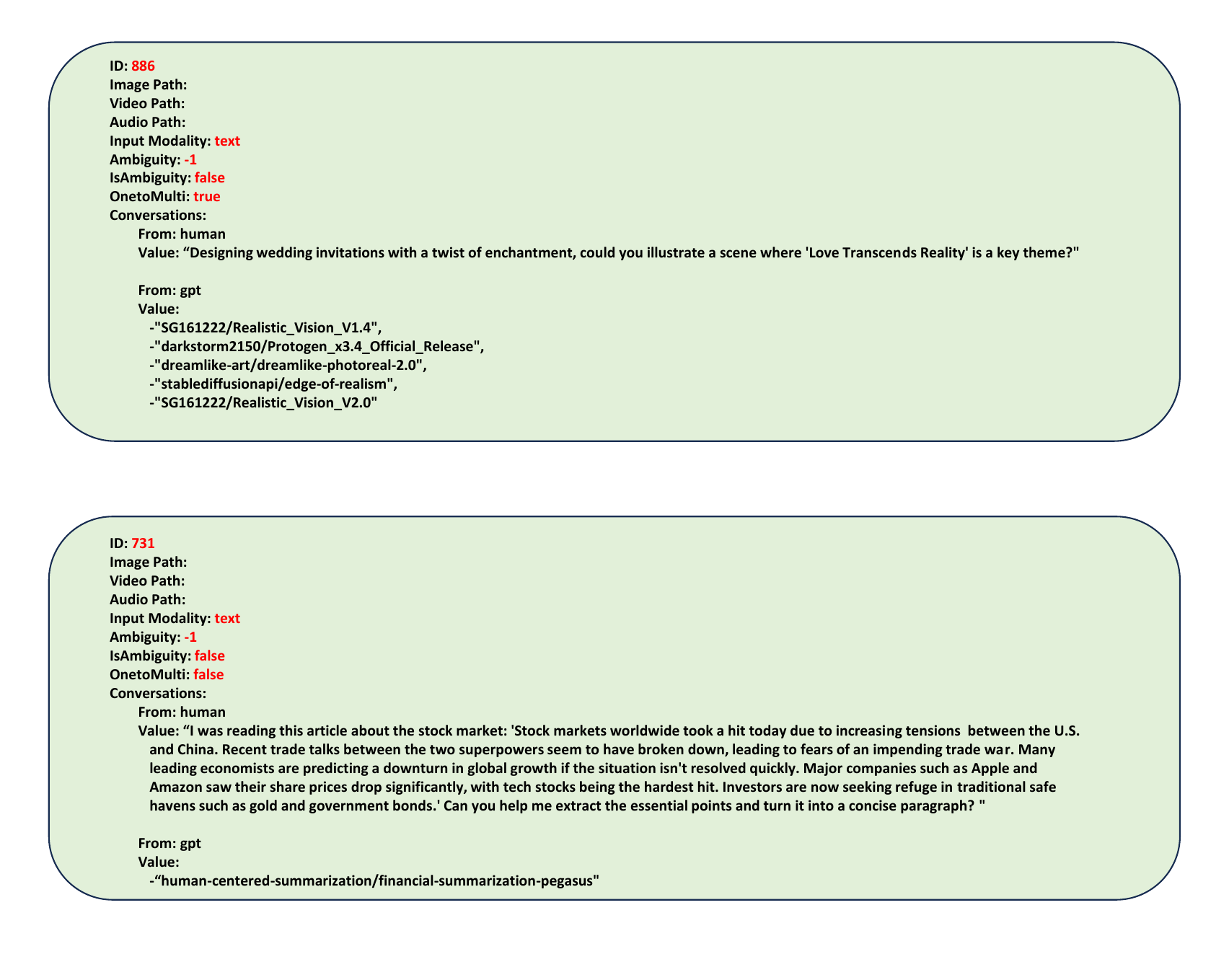}
	\caption{Instruction-Answer Pairs Visualization. (Sample 4-5)}
	\label{fig:iap2}
\end{figure*}

\begin{figure*}[htbp]
	\centering
	\includegraphics[width=\linewidth]{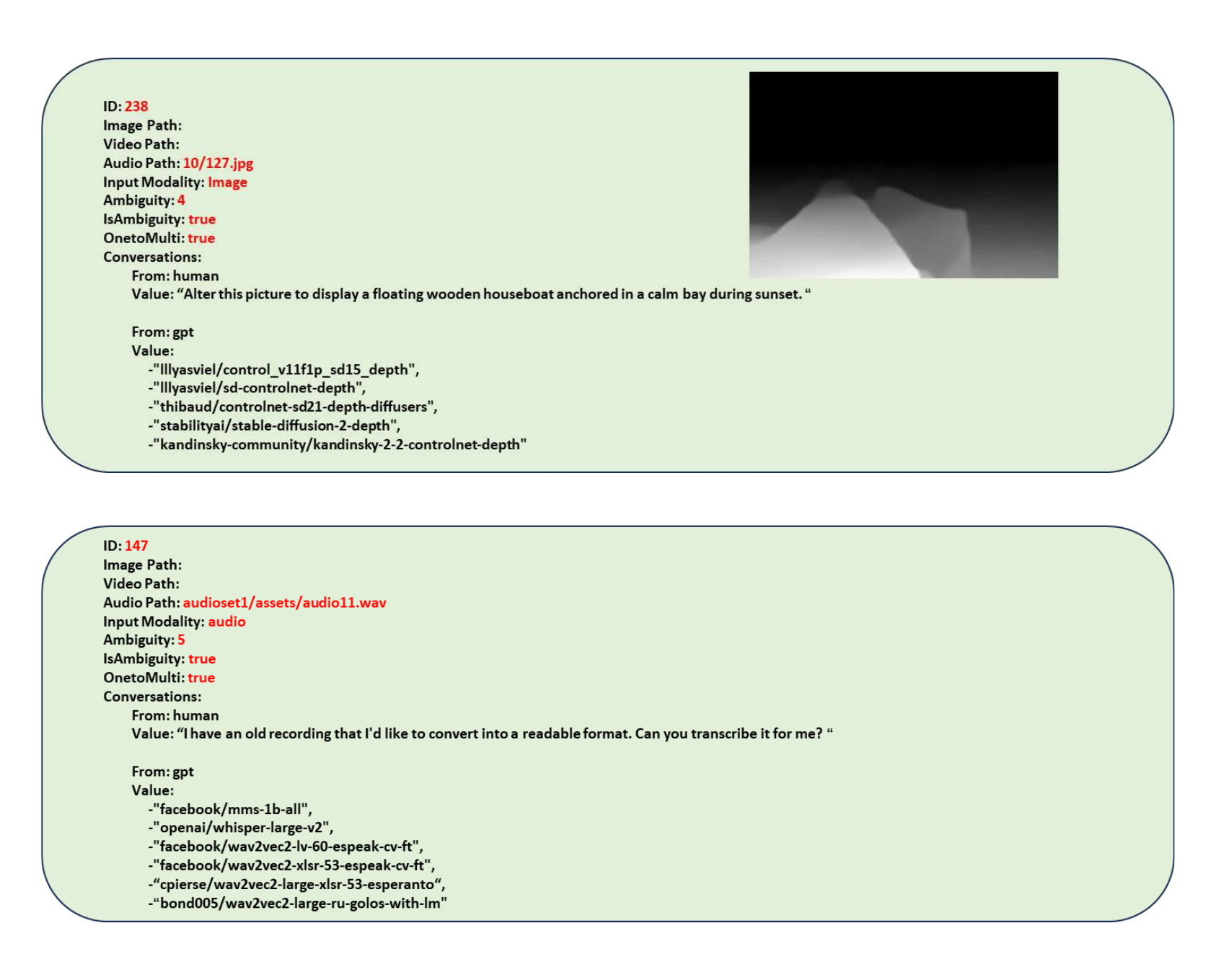}
	\caption{Instruction-Answer Pairs Visualization. (Sample 6-7)}
	\label{fig:iap3}
\end{figure*}

In Figure \ref{fig:iap1}-\ref{fig:iap3}, we offer 7 cases sampled from the testing set. When we train our model, we input the conversation between humans and GPT in a conversation format, and the other attributes help us divide them into different sub-testing sets.

\subsection{Agent Output Visualization}
\begin{figure*}[htbp]
	\centering
	\includegraphics[width=\linewidth]{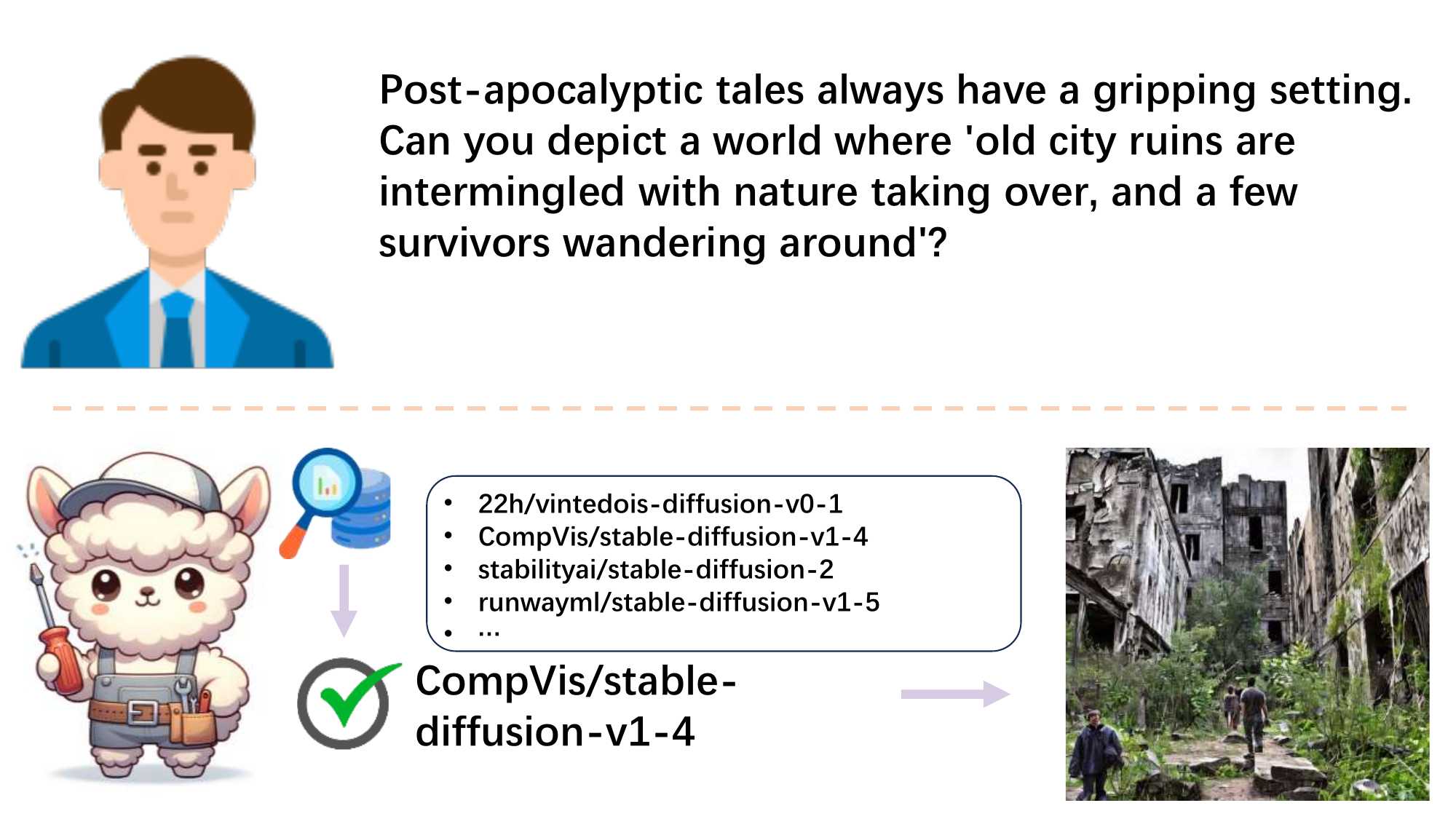}
	\caption{Agent Output Visualization. (Sample 1)}
	\label{fig:agent1}
\end{figure*}

\begin{figure*}[htbp]
	\centering
	\includegraphics[width=\linewidth]{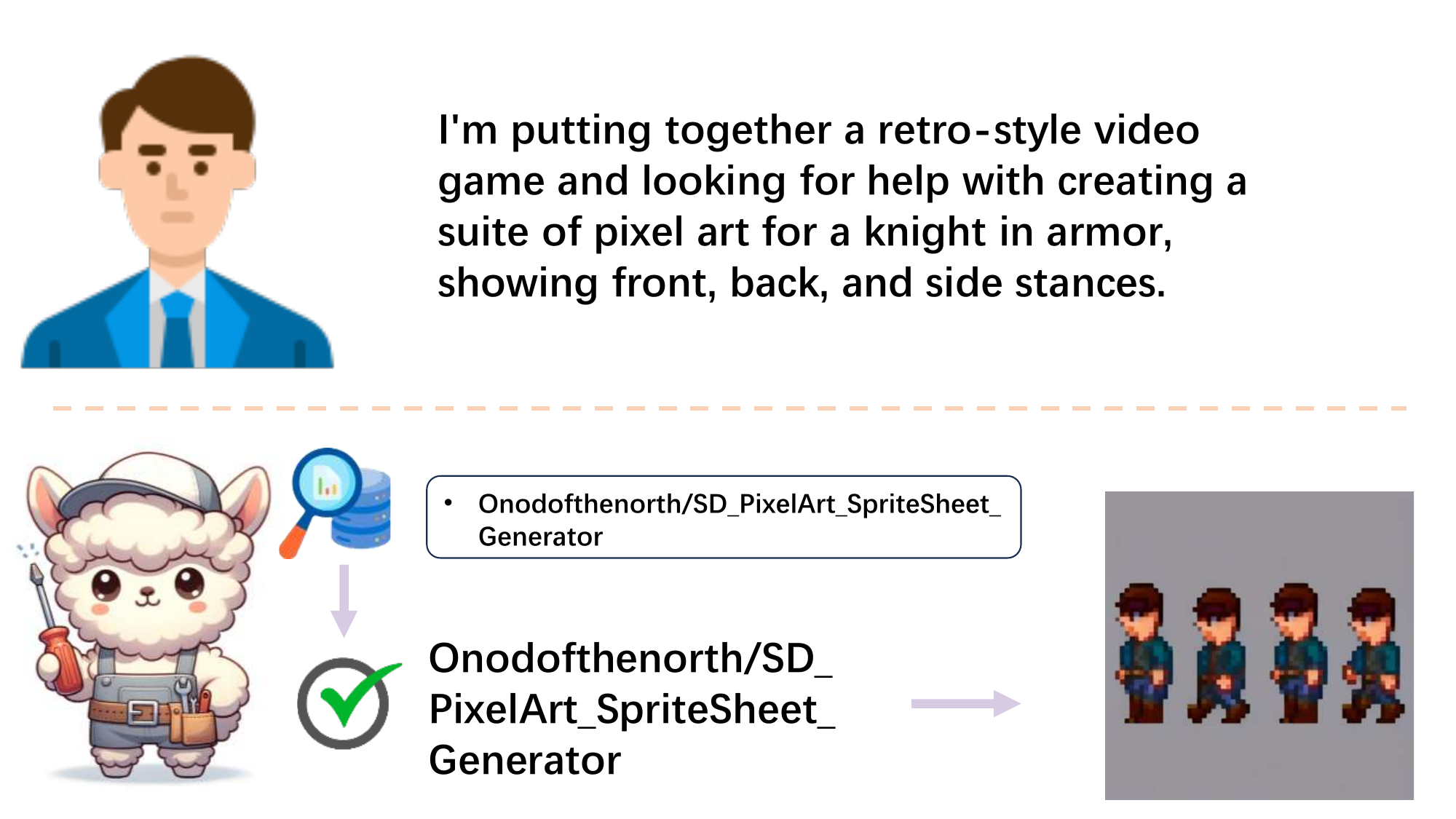}
	\caption{Agent Output Visualization. (Sample 2)}
	\label{fig:agent2}
\end{figure*}

\begin{figure*}[htbp]
	\centering
	\includegraphics[width=\linewidth]{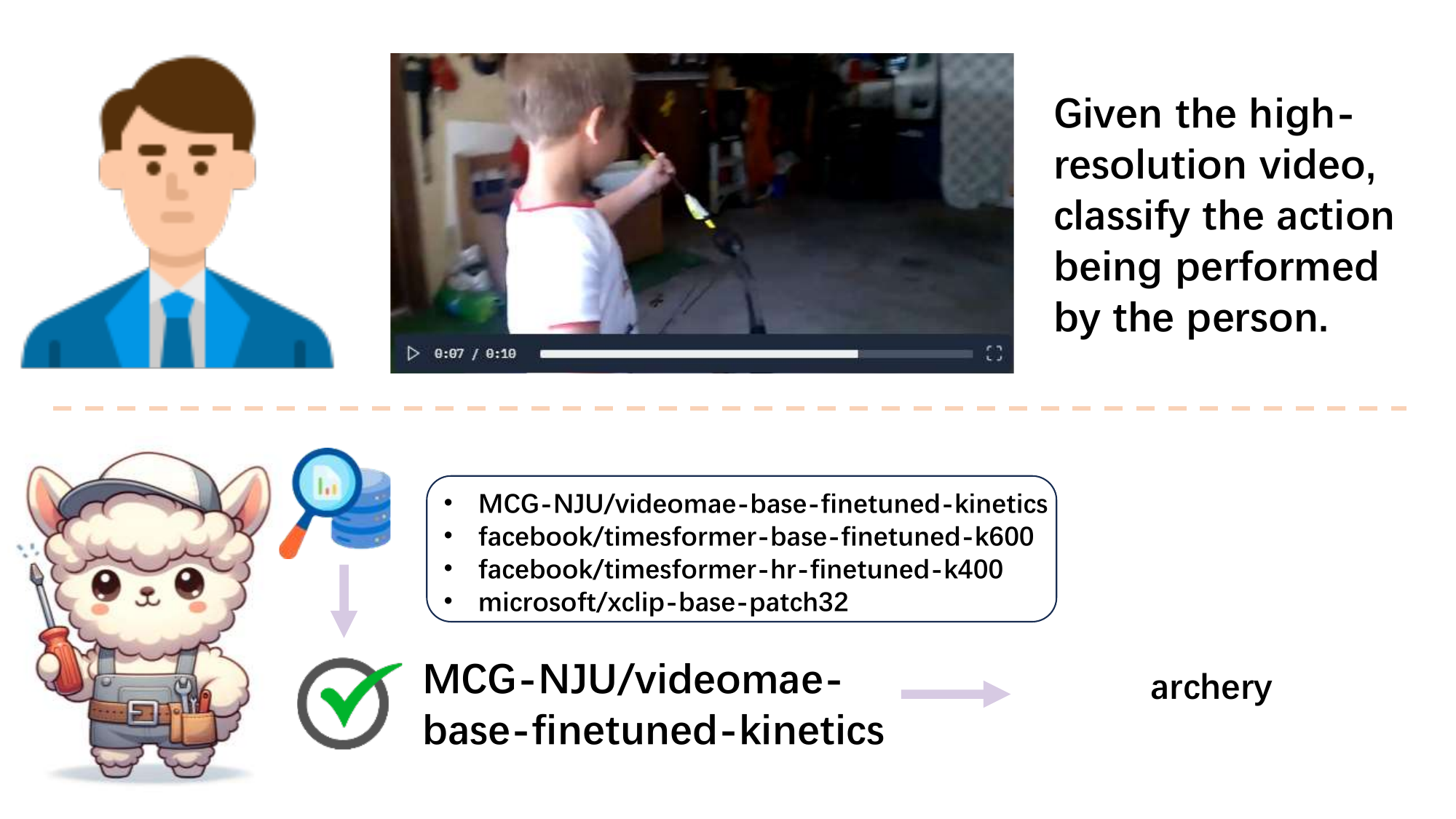}
	\caption{Agent Output Visualization. (Sample 3)}
	\label{fig:agent3}
\end{figure*}

\begin{figure*}[htbp]
	\centering
	\includegraphics[width=\linewidth]{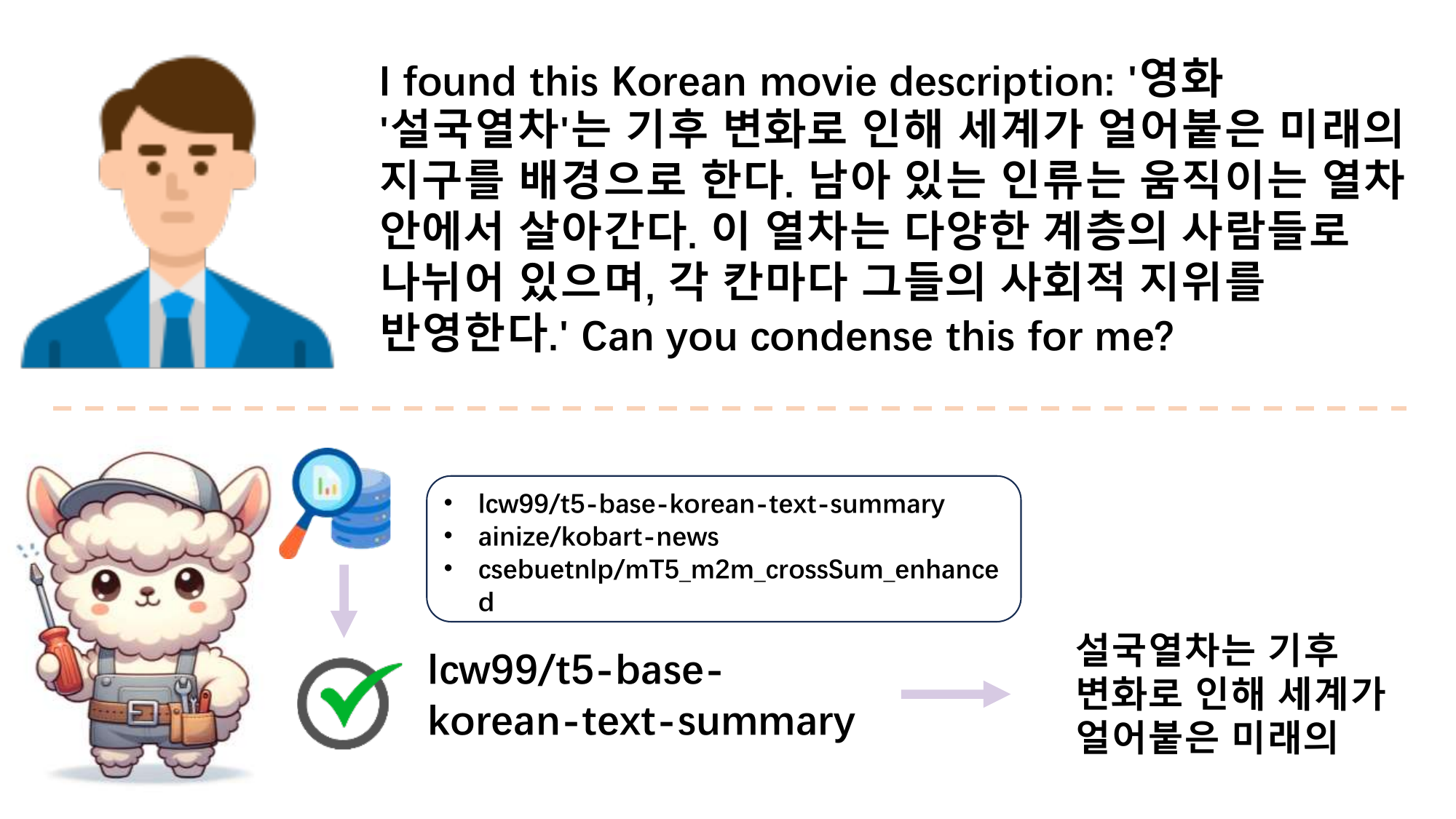}
	\caption{Agent Output Visualization. (Sample 4)}
	\label{fig:agent4}
\end{figure*}

We show four actual cases using our MLLM-Tool in Figure \ref{fig:agent1}-\ref{fig:agent4}. Our system supports inputs from four modalities and leverages the predicted API to execute, finally getting customers' expected results. 

\section{Categorization System}
We illustrate our proposed categorization system in Table \ref{tab:category_A},\ref{tab:category_B},\ref{tab:category_C}. The system is two-level hierarchical. The coarse-level task setting mainly refers to the classification of HuggingFace, while the fine-grained subtask is based on the different task descriptions.

\begin{table*}[ht]
\small
  \centering
  \resizebox{\linewidth}{!}{
    \begin{tabular}{ccc}
    \toprule
    Task & Subtask & Api Names \\
    \midrule
    \multirow{7}[2]{*}{Audio Classification} & Event Recognition & MIT/ast-finetuned-audioset-10-10-0.4593 \\
          & Command Recognition & MIT/ast-finetuned-speech-commands-v2 \\
          & Spoken Language Identification & TalTechNLP/voxlingua107-epaca-tdnn \\
          & Speaker Verification & anton-l/wav2vec2-base-superb-sv \\
          & Emotion Recognition & audeering/wav2vec2-large-robust-12-ft-emotion-msp-dim \\
          & Gender Recognition & m3hrdadfi/hubert-base-persian-speech-gender-recognition \\
          & Keyword Spotting & superb/hubert-base-superb-ks \\
    \midrule
    \multirow{6}[2]{*}{Audio-to-Audio} & Single-Channel Speech Enhancement & JorisCos/ConvTasNet\_Libri1Mix\_enhsingle\_16k \\
          & Speech-to-Speech Translation & facebook/xm\_transformer\_unity\_en-hk \\
          & Voice Conversion & microsoft/speecht5\_vc \\
          & Seperate Clean & mpariente/DPRNNTasNet-ks2\_WHAM\_sepclean \\
          & Audio Source Separation & speechbrain/sepformer-libri3mix \\
          & Pop-to-Piano & sweetcocoa/pop2piano \\
    \midrule
    Feature Extraction & Audio Feature Extraction & LeBenchmark/wav2vec2-FR-7K-large \\
    \midrule
    Audio-to-Text & Automatic Speech Recognition & Harveenchadha/vakyansh-wav2vec2-hindi-him-4200 \\
    \midrule
    \multirow{2}[2]{*}{Voice Activity Detection} & Speaker Segmentation & philschmid/pyannote-segmentation \\
          & Speaker Diarization & philschmid/pyannote-speaker-diarization-endpoint \\
    \midrule
    Depth Estimation & Depth Estimation & Intel/dpt-hybrid-midas \\
    \midrule
    \multirow{3}[2]{*}{Image Classification} & General Image Classification & facebook/convnextv2-tiny-1k-224 \\
          & Specific Type Image Classification & Neto71/sea\_mammals \\
          & Style Classifier & playrobin/furniture-styles \\
    \midrule
    \multirow{5}[2]{*}{Image Segmentation} & Text Guided Image Segmentation & CIDAS/clipseg-rd64-refined \\
          & Semantic Segmentation & apple/deeplabv3-mobilevit-small \\
          & Panoptic Segmentation & facebook/detr-resnet-50-panoptic \\
          & Instance Segmentation & facebook/mask2former-swin-small-coco-instance \\
          & Zero-shot Segmentation & facebook/sam-vit-huge \\
    \midrule
    \multirow{6}[2]{*}{Image-to-Image} & Image Super-Resolution & caidas/swin2SR-classical-sr-x2-64 \\
          & Image Variations & lambdalabs/sd-image-variations-diffusers \\
          & 2D-to-3D & openai/shap-e-img2img \\
          & Image Deblurring & google/maxim-s3-deblurring-gopro \\
          & Image Dehazing & google/maxim-s2-dehazing-sots-outdoor \\
          & Image Deraining & google/maxim-s2-deraining-rain13k \\
     \midrule
    Feaure Extraction & Image Feature Extraction & BridgeTower/bridgetower-base \\
    \midrule
    \multirow{3}[2]{*}{Object Detection} & Object Detection & SenseTime/deformable-detr \\
          & Table Detection & TahaDouaji/detr-doc-table-detection \\
          & Text-conditioned Object Detection & google/owlvit-base-patch32 \\
    \midrule
    \multirow{5}[2]{*}{Visual Question Answering} & Visual Question Answering & Salesforce/blip-vqa-base \\
          & Chart Question Answering & google/matcha-chartqa \\
          & Diagram Question Answering & google/pix2struct-ai2d-base \\
          & Document Question Answering & google/pix2struct-docvqa-base \\
          & Infographics Question Answering & google/pix2struct-infographics-vqa-large \\
    \bottomrule
    \end{tabular}}%
      \caption{Our proposed categorization system (Part 1). The first and second columns are the two levels of the categories for each task. The third column provides an example of the candidate models.}
      \label{tab:category_A}%
\end{table*}%

\begin{table*}[htbp]
\footnotesize
  \centering
  \resizebox{\linewidth}{!}{
    \begin{tabular}{ccc}
    \toprule
    Task & Subtask & Api Names \\
    \midrule
    \multirow{6}[2]{*}{Image-to-Text} & Conditional Image Caption & Salesforce/blip2-flan-t5-xl \\
          & Optical Character Recognition & alibaba-damo/mgp-str-base \\
          & Image Caption & bipin/image-caption-generator \\
          & Document Parsing & naver-clova-ix/donut-base-finetuned-cord-v2 \\
          & Tag Generation & SmilingWolf/wd-v1-4-convnextv2-tagger-v2 \\
          & Chart-to-Table & google/deplot \\
    \midrule
    Zero-Shot Image Classification & Zero-Shot Image Classification & laion/CLIP-ViT-B-16-laion2B-s34B-b88K \\
    \midrule
    \multirow{17}[2]{*}{Multimodal-to-Image} & Face-Detection-Conditioned Prompt-guided Image Generation & CrucibleAI/ControlNetMediaPipeFace \\
          & Prompt-guided Cartoonization & instruction-tuning-sd/cartoonizer \\
          & Brightness-Conditioned Prompt-guided Image Generation & ioclab/control\_v1p\_sd15\_brightness \\
          & Prompt-guided Pixel-to-Pixel Image Editing & lllyasviel/control\_v11e\_sd15\_ip2p \\
          & Shuffle-Conditioned Prompt-guided Image Generation & lllyasviel/control\_v11e\_sd15\_shuffle \\
          & Tiled-Conditioned Prompt-guided Image Generation & lllyasviel/control\_v11f1e\_sd15\_tile \\
          & Depth-Conditioned Prompt-guided Image Generation & lllyasviel/control\_v11f1p\_sd15\_depth \\
          & Canny-Edges-Conditioned Prompt-guided Image Generation & lllyasviel/control\_v11p\_sd15\_canny \\
          & Line-Art-Conditioned Prompt-guided Image Generation & lllyasviel/control\_v11p\_sd15\_lineart \\
          & MLSD-Conditioned Prompt-guided Image Generation & lllyasviel/control\_v11p\_sd15\_mlsd \\
          & Normal-Conditioned Prompt-guided Image Generation & lllyasviel/control\_v11p\_sd15\_normalbae \\
          & Openpose-Conditioned Prompt-guided Image Generation & lllyasviel/control\_v11p\_sd15\_openpose \\
          & Scribble-Conditioned Prompt-guided Image Generation & lllyasviel/control\_v11p\_sd15\_scribble \\
          & Segment-Conditioned Prompt-guided Image Generation & lllyasviel/control\_v11p\_sd15\_seg \\
          & HED-Boundary-Conditioned Prompt-guided Image Generation & lllyasviel/control\_v11p\_sd15\_softedge \\
          & Anime-Line-Art-Conditioned Prompt-guided Image Generation & lllyasviel/control\_v11p\_sd15s2\_lineart\_anime \\
          & Prompt-guided Image Variations & stabilityai/stable-diffusion-2-1-unclip \\
    \midrule
    \multirow{8}[2]{*}{Text Classification} & Sentiment Analysis & DTAI-KULeuven/robbert-v2-dutch-sentiment \\
          & Emotion Analysis & MilaNLProc/feel-it-italian-emotion \\
          & Offensive Language Detection & Hate-speech-CNERG/bert-base-uncased-hatexplain \\
          & Topic Classification & MaartenGr/BERTopic\_Wikipedia \\
          & Toxicity Analysis & OpenAssistant/reward-model-deberta-v3-large-v2 \\
          & Sensitive Text Detection & apanc/russian-inappropriate-messages \\
          & Irony Detection & cardiffnlp/twitter-roberta-base-irony \\
          & Bias Detection & cffl/bert-base-styleclassification-subjective-neutral \\
    \midrule
    Feature Extraction & Text Feature Extraction & facebook/bart-large \\
    \midrule
    Fill-Mask & Fill-Mask & CLTL/MedRoBERTa.nl \\
    \midrule
    \multirow{4}[2]{*}{Text Generation} & Text Generation & EleutherAI/gpt-j-6b \\
          & Prompt Generation & FredZhang7/anime-anything-promptgen-v2 \\
          & Code Generation & NumbersStation/nsql-350M \\
          & Specific Text Genre Generation & uer/gpt2-chinese-poem \\
    \bottomrule
    \end{tabular}}%
      \caption{Our proposed categorization system(Part 2). The first and second columns are the two levels of the categories for each task. The third column provides an example of the candidate models.}
      \label{tab:category_B}
\end{table*}%

\begin{table*}[htbp]
\small
  \centering
  \resizebox{\linewidth}{!}{
    \begin{tabular}{ccc}
    \toprule
    Task & Subtask & Api Names \\
    \midrule
    \multirow{4}[1]{*}{Text Generation} & Text Generation & EleutherAI/gpt-j-6b \\
          & Prompt Generation & FredZhang7/anime-anything-promptgen-v2 \\
          & Code Generation & NumbersStation/nsql-350M \\
          & Specific Text Genre Generation & uer/gpt2-chinese-poem \\
    \midrule
    \multirow{2}[2]{*}{Question Answering} & Extractive Question Answering & CATIE-AQ/QAmembert \\
          & Open Domain Question Answering & facebook/dpr-ctx\_encoder-single-nq-base \\
    \midrule
    Sentence Similarity & Sentence Similarity & intfloat/e5-small-v2 \\
    \midrule
    \multirow{3}[2]{*}{Summarization} & Summarization & IDEA-CCNL/Randeng-Pegasus-523M-Summary-Chinese \\
          & Title Generation & JulesBelveze/t5-small-headline-generator \\
          & Keyword Generation & Voicelab/vlt5-base-keywords \\
    \midrule
    \multirow{5}[2]{*}{Text-to-Image} & Text to General Style Image & runwayml/stable-diffusion-v1-5 \\
          & Text to RGBD Image & Intel/ldm3d \\
          & Text to Specific Style Image & nitrosocke/spider-verse-diffusion \\
          & Text to 3D Image & openai/shap-e \\
          & Text to Spectrogram Image & riffusion/riffusion-model-v1 \\
    \midrule
    Text-to-Speech & Text-to-Speech & Voicemod/fastspeech2-en-male1 \\
    \midrule
    Text-to-Video & Text-to-Video & damo-vilab/text-to-video-ms-1.7b \\
    \midrule
    \multirow{15}[2]{*}{Text-to-Text} & General Text Generation & 1-800-BAD-CODE/xlm-roberta\_punctuation\_fullstop\_truecase \\
          & Sentence Correction & KES/T5-KES \\
          & Generative Question Answering & MaRiOrOsSi/t5-base-finetuned-question-answering \\
          & Question Generation & allenai/t5-small-squad2-question-generation \\
          & Paraphraser & cointegrated/rut5-base-paraphraser \\
          & Recipe Generation & flax-community/t5-recipe-generation \\
          & Question Answering Generation & google/t5-small-ssm-nq \\
          & Text Revision & grammarly/coedit-large \\
          & Relation Extraction & ibm/knowgl-large \\
          & Code Review & microsoft/codereviewer \\
          & Commonsense Reasoning & mrm8488/t5-base-finetuned-common\_gen \\
          & Span Sentiment Extraction & mrm8488/t5-base-finetuned-span-sentiment-extraction \\
          & Distractor & potsawee/t5-large-generation-race-Distractor \\
          & Detoxification & s-nlp/bart-base-detox \\
          & Symbolic Music Generation & sander-wood/text-to-music \\
    \midrule
    Translation & Translation & Babelscape/mrebel-large \\
    \midrule
    Zero-Shot Classification & Zero-Shot Classification & MoritzLaurer/DeBERTa-v3-base-mnli-fever-anli \\
    \midrule
    Video Classification & Human action recognition video classification & MCG-NJU/videomae-base-finetuned-kinetics \\
    \midrule
    \multirow{2}[2]{*}{Video-to-Text} & Video Question Answering & kpyu/video-blip-flan-t5-xl-ego4d \\
          & Video caption & Neleac/timesformer-gpt2-video-captioning \\
    \midrule
    Feature Extraction & Video Feature Extraction & deepmind/multimodal-perceiver \\
    \bottomrule
    \end{tabular}}%
      \caption{Our proposed categorization system(Part 3). The first and second columns are the two levels of the categories for each task. The third column provides an example of the candidate models.}
    \label{tab:category_C}
\end{table*}%


\newpage
\clearpage
\clearpage

{\small
\bibliographystyle{ieee_fullname}
\bibliography{main}
}

\end{document}